\documentclass[journal,twoside,web]{ieeecolor}
\usepackage{tmi}
\usepackage{cite}
\usepackage{amsmath,amssymb,amsfonts}
\usepackage{algorithmic}
\usepackage{comment}
\usepackage{graphicx}
\usepackage{textcomp}
\usepackage{subfigure}
\usepackage{hyperref}
\usepackage{xcolor}
\usepackage{multirow}
\usepackage{booktabs} 

\usepackage[switch]{lineno}

\AtBeginDocument{
  
}
\definecolor{mygreen}{rgb}{0,0.5,0}
\def\BibTeX{{\rm B\kern-.05em{\sc i\kern-.025em b}\kern-.08em
    T\kern-.1667em\lower.7ex\hbox{E}\kern-.125emX}}
\begin{document}
\title{MetaSSL: A General Heterogeneous Loss for Semi-Supervised Medical Image Segmentation}
\author{Weiren Zhao, Lanfeng Zhong, Xin Liao, Wenjun Liao, Sichuan Zhang, Shaoting Zhang, Guotai Wang \IEEEmembership{}
\thanks{This work was supported by the National Natural Science Foundation of China (62271115). Corresponding author: G. Wang (guotai.wang@uestc.edu.cn)}
\thanks{Weiren Zhao, Lanfeng Zhong, Shaoting Zhang and Guotai Wang are with the School of Mechanical and Electrical Engineering, University of Electronic Science and Technology of China, Chengdu, 611731, China.}
\thanks{Shaoting Zhang and Guotai Wang are also with Shanghai Artificial Intelligence Laboratory, Shanghai 200030, China.}
\thanks{Wenjun Liao and Sichuan Zhang are Department of Radiation Oncology, Sichuan Cancer Hospital and Institute, University of Electronic Science and Technology of China, Chengdu, China}
\thanks{Xin Liao is with Department of Pathology, West China Second University Hospital, Sichuan University, Chengdu 610041, China.}
\thanks{This work has been accepted by IEEE TMI. Copyright of this paper has been transferred to IEEE. }
}

\maketitle

\begin{abstract}
Semi-Supervised Learning (SSL) is important for reducing the annotation cost for medical image segmentation models. State-of-the-art SSL methods such as Mean Teacher, FixMatch and Cross Pseudo Supervision (CPS) are mainly based on consistency regularization or pseudo-label supervision between a reference prediction and a supervised prediction. Despite the effectiveness, they have overlooked the potential noise in the labeled data, and mainly focus on strategies to generate the reference prediction, while ignoring the heterogeneous values of different unlabeled pixels.  
We argue that effectively mining the rich information contained by the two predictions in the loss function, instead of the specific strategy to obtain a reference prediction, is more essential for SSL, and propose a universal framework \textbf{MetaSSL} based on a spatially heterogeneous loss that assigns different weights to pixels by simultaneously leveraging the uncertainty and consistency information between the reference and supervised predictions. Specifically, we split the predictions on unlabeled data into four regions with decreasing weights in the loss: Unanimous and Confident (UC), Unanimous and Suspicious (US), Discrepant and Confident (DC), and Discrepant and Suspicious (DS), where an adaptive threshold is proposed to distinguish confident predictions from suspicious ones.  
The heterogeneous loss is also applied to labeled images for robust learning considering the potential annotation noise. Our method is plug-and-play and general to most existing SSL methods. The experimental results showed that it improved the segmentation performance significantly when  integrated with existing SSL frameworks on different datasets. 
Code is available at \href{https://github.com/HiLab-git/MetaSSL}{https://github.com/HiLab-git/MetaSSL}.
\end{abstract}


\section{Introduction}
\label{sec:introduction}


\begin{figure}
    \centering
    \includegraphics[width=1\linewidth]{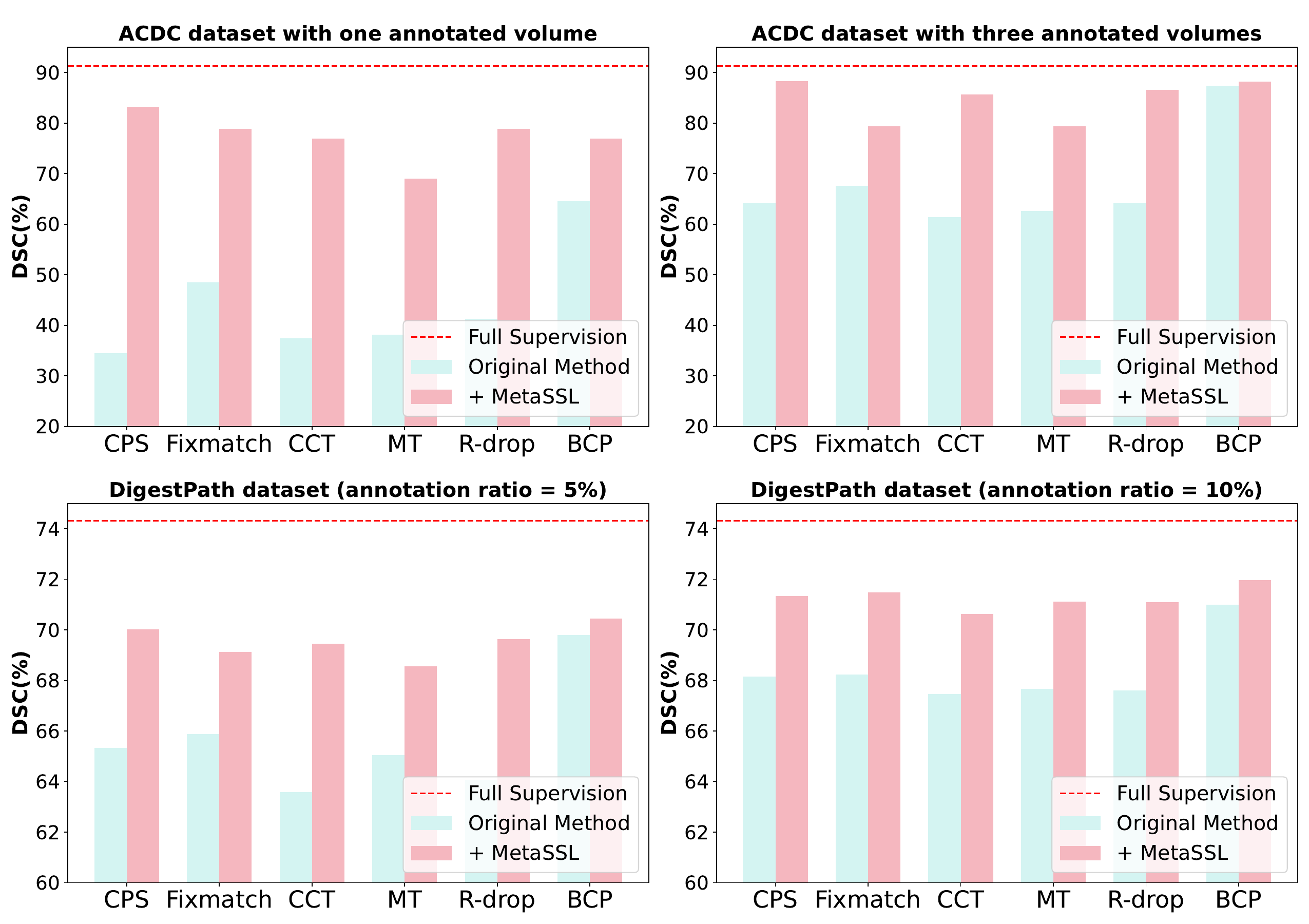}
    \caption{Comparison of Dice Similarity Coefficient (DSC) obtained by our MetaSSL on two different datasets with different annotation ratios when applied to  several existing SSL methods, including CPS~\cite{chen2021semi}, FixMatch~\cite{sohn2020fixmatch}, CCT~\cite{ouali2020semi}, MT~\cite{tarvainen2017mean}, R-Drop~\cite{liang2021r} and BCP~\cite{bai2023bidirectional}.}
    \label{fig:different_backbone}
\end{figure}

\IEEEPARstart{D}{eep}  learning has made significant progress in achieving high performance on medical image segmentation  that is a fundamental step for quantitative evaluation of diseases, 3D modeling of anatomical structures, and treatment planning and guidance~\cite{suganyadevi2022review,razzak2018deep, xie2019recurrent}. However, training a high-performance segmentation model requires dense annotations of a large-scale dataset that are expansive, time-consuming and labor-intensive to obtain, as accurate dense annotations need to be collected from experienced radiologists who are hardly accessible. To handle this problem, annotation-efficient learning has gained increasing attentions to unleash the potential of deep learning in more widespread clinical applications~\cite{wang2022pymic}. Considering that it is often practical to obtain a small number of labeled images with more unlabeled ones, Semi-Supervised Learning (SSL) is promising for obtaining high segmentation performance while keeping a low annotation cost~\cite{wang2022pymic}.

To deal with unannotated images, most existing SSL methods generate multiple diverse predictions for the same input so that one is used as a reference prediction to supervise the others~\cite{tarvainen2017mean,sohn2020fixmatch,chen2021semi}. The strategies to use reference prediction mainly include two types: consistency regularization and pseudo-labeling, and various approaches have been proposed  to generate these multiple predictions, 
such as Mean Teacher (MT)~\cite{tarvainen2017mean} that uses a teacher's output as the reference prediction to supervise a student via a consistency loss, FixMatch~\cite{sohn2020fixmatch} that uses output from a weakly augmented version 
to supervise that from a strongly augmented version, and Cross Consistency-Training (CCT)~\cite{ouali2020semi} that 
encourages several auxiliary decoders' predictions to be consistent with a reference prediction  from the main decoder. 
Despite extensive efforts on perturbations at input~\cite{sohn2020fixmatch,yun2019cutmix,bai2023bidirectional}, feature~\cite{ouali2020semi} or network structure~\cite{Luo2022media,zhong2024pr} levels to construct multiple predictions with diversity, existing multi-prediction-based SSL methods for image segmentation have three common drawbacks: 

Firstly, they pay little attention to effectively mining the rich information contained in the reference and supervised predictions for learning, and typically use simple spatially homogeneous  loss functions for consistency regularization or pseudo-label supervision. For example, Mean Squared Error (MSE) was used by consistency regularization-based SSL methods such as Mean Teacher~\cite{tarvainen2017mean}, CutMix~\cite{yun2019cutmix}, CCT~\cite{ouali2020semi}, MixMatch~\cite{berthelot2019mixmatch} and MC-Net~\cite{wu2022mutual}, and naive Cross Entropy (CE) loss or Dice loss is used by pseudo-labeling SSL methods including FixMatch~\cite{sohn2020fixmatch}, CPS~\cite{chen2021semi}, BCP~\cite{bai2023bidirectional} and ABD~\cite{Chi2024}.
These methods often treat different pixels equally, without considering the quality of the reference prediction at different positions. Though some methods have tried to leverage uncertainly estimation to suppress the effect of unconfident regions in the reference prediction~\cite{Yu2019miccai,xia2020uncertainty,suganyadevi2022review}, they have ignored the discrepancy or consistency  between the reference and supervised predictions, which is important for more informed learning. Some other methods also combined uncertainty and consistency~\cite{Luo2022media,wu2022mutual}, but they mainly introduce  uncertainty-aware consistency regularization, without more comprehensive analysis and elaborated weighting of certain/uncertain and consistent/discrepant regions.

Secondly, due to the low contrast of medical images and intra-observer variations during annotation, the labeled data inevitably contain some noise, whereas existing SSL methods often assume that these labels are absolutely clean, which can hardly hold in practice. In addition, even for clean annotations, different pixels contribute differently to the model's convergence, i.e., some hard regions may be more informative for performance improvement~\cite{chen2023adaptive,salehi2017tversky,wang2023conflict}. The inequality of pixels in the annotated images has been ignored in most existing SSL methods, and a more elaborated loss to deal with different parts of the labeled images has a potential for further improving the performance of SSL. 

Thirdly, despite the wide use of uncertainty and consistency in existing SSL methods, the network architectures and unlabeled loss functions are designed ad hoc, such as obtaining multiple predictions by different networks~\cite{chen2021semi,tarvainen2017mean}, resolutions~\cite{Luo2022media}, decoders~\cite{ouali2020semi,wu2022mutual} or forward passes~\cite{liang2021r,Yu2019miccai}. While they share some common underlying mechanisms of leveraging multiple predictions, their generalization to different SSL frameworks is limited, and a more unified unsupervised loss for unlabeled images that is generalizable to different SSL frameworks are desirable. 


Based on the above observations, we hypothesize that an elaborated loss function to better mine the useful information and suppress noise for both unlabeled and labeled images are important for SSL methods. Therefore, we propose a spatially heterogeneous loss for SSL that assigns different weights to pixels by simultaneously leveraging the uncertainty and consistency information between the reference and supervised predictions, due to that confident regions in the reference prediction are more likely to be correct than uncertain regions~\cite{Yu2019miccai,media2022urpc}, and inconsistent parts between the reference and supervised predictions are informative for effective learning~\cite{shi2021inconsistency}. It splits the reference and supervised predictions into four regions with decreasing weights in the loss: Unanimous and Confident (UC), Unanimous and Suspicious (US), Discrepant and Confident (DC), and Discrepant and Suspicious (DS). Note that the reference could be both pseudo-labels predicted by a network for unlabeled images and manual annotations of labeled images, therefore our method is general to unlabeled and labeled images in SSL. In addition, it does not requires a specific strategy to generate the reference prediction, and is general to most existing multi-prediction-based SSL methods.  As shown in ~\autoref{fig:different_backbone}, when our method is applied to five state-of-the-art SSL methods including Mean Teacher~\cite{tarvainen2017mean}, FixMatch~\cite{sohn2020fixmatch}, CCT~\cite{ouali2020semi}, R-Drop~\cite{liang2021r} and CPS~\cite{chen2021semi}, significant improvement is obtained on different datasets and different annotation ratios. The main contributions of this work are summarized as follows:

\begin{itemize}
    \item We propose a general semi-supervised medical image segmentation method MetaSSL based on a spatially heterogeneous loss function, which partitions the reference prediction 
    into four categories with different weights based on confidence and consistency between the reference and supervised predictions, so that more valuable regions can be highlighted for more effective learning on unannotated images. 
    \item We also extend the spatially heterogeneous loss to labeled images in SSL, which helps to mine hard-to-learn regions and suppress potential noisy annotations to improve the performance of semi-supervised segmentation. 
    \item We present a self-adaptive threshold strategy that exhibits adaptability to distinguish confident predictions from uncertain ones for different target classes at different training stages, which better helps uncertainty-aware SSL. 
\end{itemize}

Our method was validated on three public medical image segmentation datasets. The experimental results showed that it outperformed 12 state-of-the-art SSL methods, and significant performance improvement was achieved when integrated into different existing multi-prediction-based SSL methods.

\section{Related works}

\begin{figure*}
	\centering
	\includegraphics[width=1\linewidth]{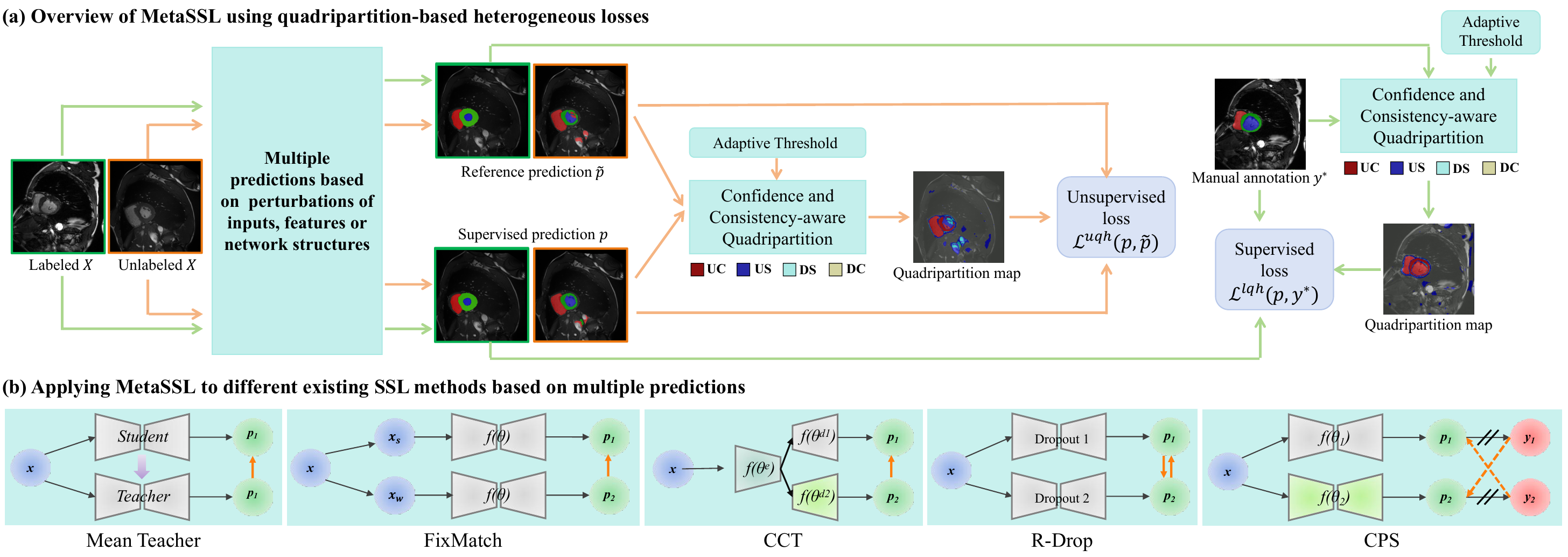}
	\caption{Overview of our MetaSSL for semi-supervised segmentation. (a) shows  the overall framework of our method that applies quadripartition-based heterogeneous losses to labeled and unlabeled images based on multiple predictions.  The quadripartition leads to four different regions: Unanimous and Confident (UC), Unanimous and Suspicious (US), Discrepant and Confident (DC), and Discrepant and Suspicious (DS), and each of them is assigned a different weight in loss calculation. The orange and green arrows show dataflow of unlabeled and labeled images, respectively. (b) Our MetaSSL can be implemented with different existing SSL methods that obtain multiple predictions for pseudo-labeling or consistency regularization. For simplicity, the brown arrows represent our quadripartition-based heterogeneous loss for unlabeled images. }
	\label{fig:ours}
\end{figure*}
\subsection{Semi-Supervised Learning for Segmentation}
To exploit the potential of unlabeled data, existing Semi-Supervised Learning (SSL) methods  mainly use two strategies: pseudo-labeling and consistency regularization. 
Among pseudo-label learning methods, self-training~\cite{yang2022st++,he2021re,bai2017semi,zhao2024semi} trains a model on labeled images to generate pseudo-labels of unlabeled images for the model itself, and iterative process is used to improve the pseudo-label quality.  
Considering that self-training may accumulate errors on unlabeled images, more recent methods~\cite{peng2020deep,fan2022ucc,zhong2024pr} including CPS~\cite{chen2021semi} and ABD~\cite{Chi2024} employ multiple models or prediction heads to generate pseudo-labels for each other, which helps to avoid the bias towards a single model or prediction head. 

On the other hand, consistency regularization methods 
manipulate the unlabeled data by introducing perturbations to obtain multiple predictions, and encourage consistency between them to achieve robustness to such perturbations. The perturbations can be implemented at the input-level~\cite{sohn2020fixmatch,berthelot2019mixmatch}, feature-level~\cite{ouali2020semi} or structure-level~\cite{luo2022semi,wu2022mutual,zhong2024pr}. 
For example, FixMatch~\cite{sohn2020fixmatch} and MixMatch~\cite{berthelot2019mixmatch} encourage predictions from different augmented versions of an input to be consistent.  R-Drop~\cite{liang2021r} and CCT~\cite{ouali2020semi} introduce feature perturbations by dropout and adding noise, and MC-Net~\cite{wu2022mutual} and CDMA~\cite{zhong2024pr} use multiple decoders with different structures to obtain diverse predictions for consistency regularization. Despite their effectiveness, most existing SSL methods use spatially homogeneous loss functions for unlabeled images, which ignores the different usefulness of different regions in the reference prediction for training the segmentation model.  


\subsection{Loss Functions for Medical Image Segmentation}
Loss functions play an important role in medical image segmentation models~\cite{ma2021loss}. For fully supervised learning segmentation, the Cross-Entropy loss~\cite{milletari2016v} and Dice loss~\cite{ronneberger2015u,sudre2017generalised} have been widely used due to their simplicity and effectiveness. However, they treat each pixel and class equally, leading to limited performance on hard cases and small target regions. To deal with this issue, hardness-aware losses~\cite{Wong2018miccai,wang2019miccai} have been proposed to  automatically assign higher weights to hard classes or pixels. Tversky Loss~\cite{salehi2017tversky}  extends Dice Loss by introducing weighting parameters to balance False Positives (FP) and False Negatives (FN)
for dealing with imbalanced datasets. Chen et al.~\cite{chen2023adaptive} proposed an adaptive region-specific loss  that adjusts the hyper-parameters of the Tversky loss automatically based on FP  and FN ratios. In addition, some distance-aware losses~\cite{Karimi2020,Kervadec2021loss} have been  proposed to reduce distance errors for highly imbalanced segmentation.
 
Despite that several loss functions have been designed for supervised segmentation, the loss function design has been rarely investigated in semi-supervised segmentation setting. Most existing SSL methods simply use Cross-Entropy loss and Dice loss for labeled images. For unlabeled images, they also simply use these loss functions for pseudo-label learning, or MSE for consistency regularization, without considering the hard regions or noisy labels  in manual annotations and pseudo-labels in SSL, which limits the performance.

\section{Methods}




Let $D = D_l \cup D_u$ denote the training set for semi-supervised learning, where $D_l$ is the set of $M$ labeled images, and $D_u$ is the set of $N$ unlabeled images, respectively. The total number of classes for the  segmentation task is denoted by $C$.  In general, $M$ is significantly larger than $N$, i.e., $M \gg N$.




As illustrated in \autoref{fig:ours}(a), our method is built on a general multi-prediction-based SSL framework, where a reference prediction given by a network or a branch is used for supervision on unlabeled images. Based on uncertainty estimation and consistency between the reference and supervised predictions, the pixels are partitioned into four regions that are weighted heterogeneously, so that more reliable and valuable regions are highlighted in the loss function. A similar weighting scheme is applied to labeled data, which suppresses noisy annotations and highlights hard regions for better learning. Our heterogeneous loss is plug-and-play, and can be applied to several existing SSL methods, as shown in~\autoref{fig:ours}(b).   

\subsection{Confidence and Consistency-aware 
Quadripartition}\label{sec:quadripartition}
We consider a SSL framework that leverages multiple predictions with diversity sourced from perturbations for consistency regularization or pseudo-label learning in semi-supervised medical image segmentation. For an unlabeled image $X$, let $\tilde{p}$ and $p$ denote the reference and supervised predictions of a model, respectively. Note that due to perturbations on the input image~\cite{sohn2020fixmatch}, intermediate features~\cite{ouali2020semi,liang2021r} or network structures~\cite{chen2021semi,luo2022semi}, $\tilde{p}$ and $p$ have some discrepancy, and the quality of $\tilde{p}$ is expected to be not lower than that of $p$~\cite{tarvainen2017mean,sohn2020fixmatch, chen2021semi},  so that  $\tilde{p}$ can be used to supervise $p$ on unlabeled images. A general unsupervised loss between $\tilde{p}$ and $p$ is denoted as $\mathcal{L}_{unsup}=L(p, \tilde{p})$, and $L(p, \tilde{p})$ is usually implemented by MSE in consistency regularization~\cite{tarvainen2017mean,sohn2020fixmatch,wu2022mutual}, or CE/Dice loss in pseudo-label learning~\cite{chen2021semi,luo2022semi,bai2023bidirectional}. 

As $\tilde{p}$ may contain some noise or hard regions, different pixels in $\tilde{p}$ contribute differently to the learning process, treating different pixels equally in $L(p, \tilde{p})$ as implemented in most existing works may limit the segmentation model's performance, and identifying the reliable and valuable region in  $\tilde{p}$ is of great importance for SSL. For example, predictions with higher uncertainty values are usually incorrect ones that should be suppressed in supervision~\cite{Luo2022media,fan2022ucc}, and predictions with a discrepancy between two networks or two branches are more valuable for mutual learning to avoid bias~\cite{shi2021inconsistency}. Therefore, it is desirable to recognize  regions in $\tilde{p}$ with different values for learning. To achieve this goal, we comprehensively leverage confidence estimation and consistency information to partition different pixels into four categories to better highlight valuable ones and suppress noisy or less useful ones.


Let $\tilde{y}$ and $y$ denote the hard segmentation map converted by argmax from $\tilde{p}$ and $p$ respectively, and we use $\tilde{p}_{\bold{x},c}$ to denote the probability of pixel $\bold{x}$ being class $c$ in the reference prediction. The entire spatial domain in $X$ is partitioned into four regions: Unanimous and Confident (UC) region $\Omega_{UC}$, Unanimous and Suspicious (US) region  $\Omega_{US}$, Discrepant and Confident (DC) region $\Omega_{DC}$ and Discrepant and Suspicious (DS) region $\Omega_{US}$, and they are defined as: 
\begin{equation}\label{eq:uc}
\Omega_{UC} = \left\{\bold{x} \mid \tilde{y}_{\bold{x}} == y_{\bold{x}} \text{ and } \max(\tilde{p}_{\bold{x},c}) > \gamma_{c^*} \right\}
\end{equation}

\begin{equation}\label{eq:us}
\Omega_{US} = \left\{\bold{x} \mid {\tilde{y}_{\bold{x}}} == y_{\bold{x}} \text{ and } \max(\tilde{p}_{\bold{x},c}) \leq \gamma_{c^*} \right\}
\end{equation}

\begin{equation}\label{eq:dc}
\Omega_{DC} = \left\{\bold{x} \mid \tilde{y}_{\bold{x}} \neq y_{\bold{x}} \text{ and } \max(\tilde{p}_{\bold{x},c}) > \gamma_{c^*} \right\}
\end{equation}

\begin{equation}\label{eq:ds}
\Omega_{DS} = \left\{\bold{x} \mid {\tilde{y}_{\bold{x}}} \neq {y}_{\bold{x}} \text{ and } \max(\tilde{p}_{\bold{x},c}) \leq \gamma_{c^*} \right\}
\end{equation}
where $\max(\tilde{p}_{\bold{x},c})$ is the maximal probability value across all the classes in $\tilde{p}_{\bold{x},c}$, and  $\gamma_{c^*}$ is a confidence threshold value for the reference prediction with $c^*=\arg\max_{c} \tilde{p}_{\bold{x},c}$. 


\subsection{Heterogeneous Loss on Unlabeled Images}\label{sec:heterogeneus_loss}
Based on the above quadriparition of pixels in the reference prediction, we then assign different weights to these regions, leading to a spatially heterogeneous loss function for unlabeled images. The weights for $\Omega_{UC}$, $\Omega_{US}$, $\Omega_{DC}$ and $\Omega_{DS}$ are denoted as  $w_{UC}$, $w_{US}$, $w_{DC}$ and $w_{DS}$, respectively. Generally, regions with unanimous results among multiple predictions is more robust and reliable than those with discrepant results~\cite{Mehrtash2020, WANG201934}, so $w_{UC}$ and $w_{US}$ are set to be higher than $w_{DC}$ and $w_{DS}$. In addition, within the region of unanimous predictions, areas where the networks exhibit confident predictions are considered to be more reliable than those with uncertain predictions. Therefore, \( w_{UC} \) should be higher than \( w_{US} \). Similarly, \( w_{DC} \) is set to higher than \( w_{DS} \). 
This leads to $ w_{UC} > w_{US} > w_{DC} > w_{DS}$.

\begin{figure}
    \centering
    \includegraphics[width=1\linewidth]{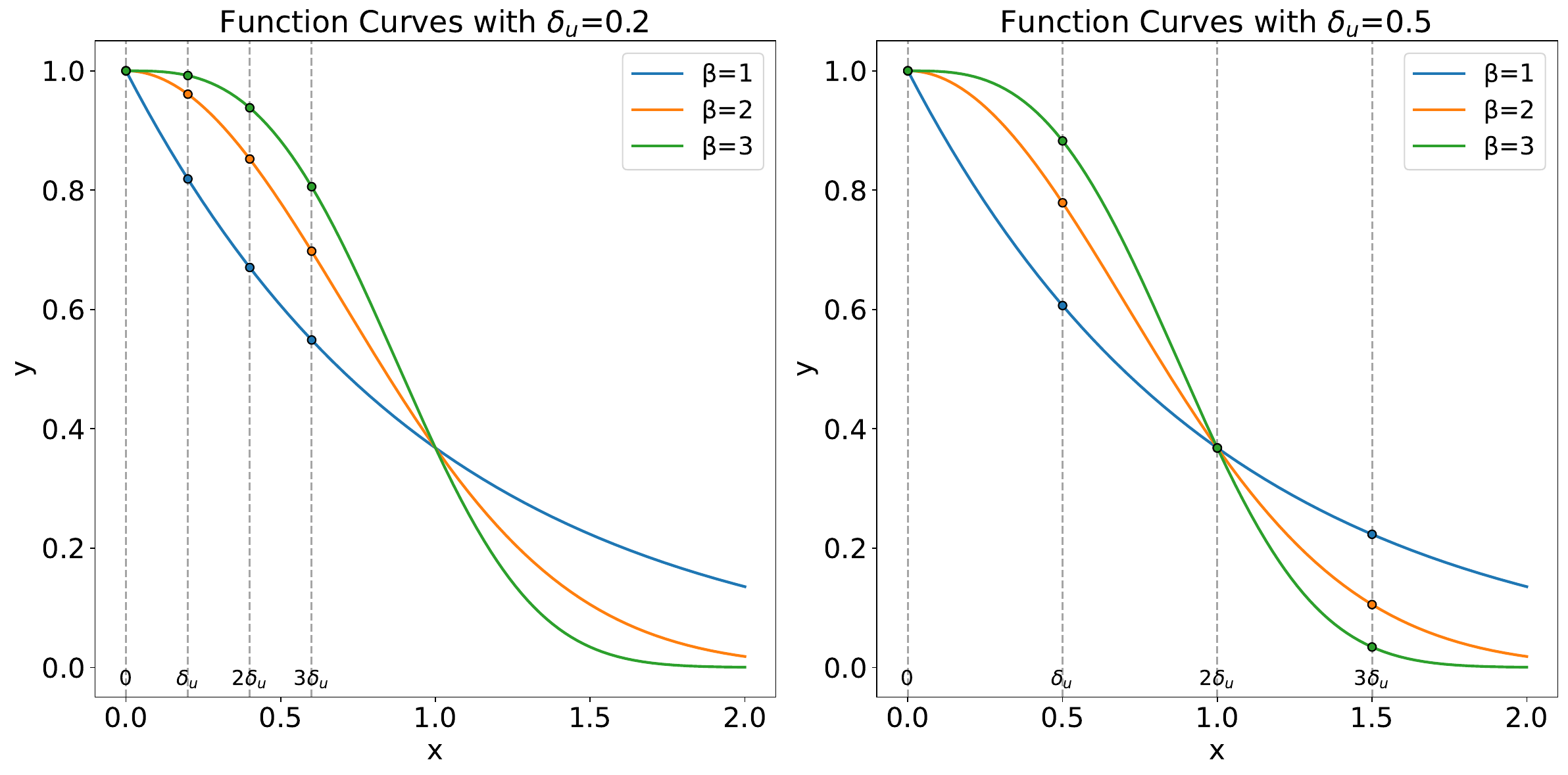}
    \caption{Illustration of quadripartition-based pixel weights under different $\beta$ and $\delta_u$ values.}
    \label{fig:m1}
\end{figure}
Let $\Omega_{all}=\{\Omega_{UC}$, $\Omega_{US}$, $\Omega_{DC}$, $\Omega_{DS}\}$ be the entire image region. $\Omega_t$ $\in$ $\Omega_{all}$ is one of these four regions, and the corresponding weight is denoted as $w_t$. The basic unsupervised loss function $L(p, \tilde{p})$ is extended to a spatially weighted version according to   $\Omega_t$. For example, for CE loss with pseudo labels on an unlabeled image, the quadripartition-based heterogeneous loss is: 
\begin{equation}\label{eq:L_unsup}
 \mathcal{L}^{uqh}(p, \tilde{p}) = - \frac{1}{Z} \sum_{\Omega_t \in \Omega_{all}}\sum_{\bold{x} \in \Omega_t} w_t\sum_{c=0}^{C-1} \tilde{y}_{\bold{x},c} \log p_{\bold{x},c}
\end{equation}
where $Z=\sum_{\Omega_t\in \Omega_{all}}w_t|\Omega_t|$  is a normalization factor. $p_{\bold{x},c}$ means the probability of class $c$ at pixel $\bold{x}$ in the prediction, and $\tilde{y}_{\bold{x},c}$ is the corresponding value in the argmax version of the reference prediction $\tilde{p}$.

To avoid tuning the value of the four weights respectively and reduce hyper-parameters, we consider a monotonically decreasing function $\phi(u)$ to generate $w_{UC}$, $w_{US}$, $w_{DC}$, and $w_{DS}$  in a descending order. Let $\delta_u$ denote the interval along the input dimension of $\phi(u)$ for unlabeled images, the value of $w_{UC}$, $w_{US}$, $w_{DC}$, and $w_{DS}$ are set as $\phi(0)$, $\phi(\delta_u)$, $\phi(2\delta_u)$ and $\phi(3\delta_u)$, respectively.
As shown in \autoref{fig:m1}, we consider a generalized Gaussian function with an input range of [0, 2.0]:

\begin{equation}\label{eq:phi}
\phi(u) = e^{-u^\beta} 
\end{equation}

By tuning the value of $\beta$ and $\delta_u$, we can control the values of the four weights. Specifically, the $\beta$ value will affect the range of the weights. For example, when $u \in$ [0, 2], the range of $\phi(u)$ will be (0.137, 1.0), (0.019, 1.0) and (3.78$\times 10^{-4}$, 1.0) for $\beta$ = 1, 2 and 3, respectively. The value of $\delta_u$ will further control the relative value of the four weights. For example, when $\beta = 3.0$, setting $\delta_u$ = 0.5 will make the value of $w_{UC}$, $w_{US}$, $w_{DC}$, and $w_{DS}$ be  1.0, 0.882, 0.368 and 0.034 respectively, leading to $w_{UC}$ and $w_{US}$ being much higher than $w_{DC}$  and $w_{DS}$. In contrast, when $\delta_u$ = 0.2, the value of $w_{UC}$, $w_{US}$, $w_{DC}$, and $w_{DS}$ will be  1.0, 0.992, 0.938 and 0.806 respectively, leading these weights to be relatively closer.

\subsection{Dealing with Noise in Labeled Images}

In labeled images, the annotation may contain noise around the boundary region and small targets when the contrast is low or the annotator delineates the boundary roughly to save time. Due to the small amount of annotated images in SSL setting, standard supervised loss on these images will easily lead to over-fitting to the noise, which limits the model's performance. To deal with this problem, we further extend the heterogeneous loss to labeled images for noise-robust learning. 

Following the multi-prediction framework in Section~\ref{sec:quadripartition}, we use $p$ and $\tilde{p}$ to denote two predictions for the same training image under perturbations in input, feature or network structure. Let $y^*$ denote the manual annotation of the corresponding image. Note that $p$ is supervised by $y^*$ here, which is different from  Section~\ref{sec:quadripartition} where $p$ is supervised by $\tilde{p}$. As $y^*$ itself does not contain confidence or error information, we leverage the discrepancy between $\tilde{p}$ and $y^*$  and the confidence of  $\tilde{p}$ to estimate reliable and unreliable regions in $y^*$. 

Following Eq.~\eqref{eq:uc} to Eq.~\eqref{eq:ds}, to supervise $p$, we also split $y^*$ into four regions ($\Omega_{UC}$, $\Omega_{US}$, $\Omega_{DC}$ and $\Omega_{DS}$) according to the discrepancy between $y^*$ 
and $\tilde{y} = \text{argmax} (\tilde{p}) $, and the confidence of $\tilde{p}$. Generally, we assume that $\Omega_{UC}$ and $\Omega_{US}$ are more reliable than  $\Omega_{DC}$ and $\Omega_{DS}$, as regions with discrepant $y^*$ and $\tilde{y}$ values are more likely to contain noisy annotations. As a result, $w_{UC}$ and $w_{US}$ are set  to be larger than $w_{DC}$ and $w_{DS}$, and $w_{UC}$ is set to be larger than $w_{US}$ to emphasize the confident part in the unanimous regions. In contrast, in the discrepant regions,  a confident $\tilde{p}$ indicates that the corresponding $y^*$ is likely to be noisy, and we set  $w_{DC}$ to be lower than $w_{DS}$, leading to a $w_{UC} > w_{US} > w_{DS} > w_{DC}$  for labeled images.  Their values are also generated by the generalized Gaussian function defined in Eq.~\eqref{eq:phi}, and $w_{UC}$, $w_{US}$, $w_{DS}$ and $w_{DC}$ are set as  $\phi(0)$, $\phi(\delta_l)$, $\phi(2\delta_l)$ and $\phi(3\delta_l)$, respectively, where $\delta_l$ is a hyper-parameter. 
Similarly to Section~\ref{sec:heterogeneus_loss}, we also extend the supervised loss on labeled images to a spatially weighted version. For example, a quadripartition-based heterogeneous cross-entropy loss between $p$ and $y^*$ is:
\begin{equation}\label{eq:l_ce_l}
 \mathcal{L}_{ce}^{lqh}(p, y^*) = - \frac{1}{Z} \sum_{\Omega_t \in \Omega_{all}}\sum_{\bold{x} \in \Omega_t} w_t\sum_{c=0}^{C-1} y^*_{\bold{x},c} \log p_{\bold{x},c}
\end{equation}
where $Z=\sum_{\Omega_t\in \Omega_{all}}w_t|\Omega_t|$ is a normalization factor, and $w_t$ is the weight of $\Omega_t \in \{\Omega_{UC}, \Omega_{US}, \Omega_{DC}, \Omega_{DS}\}$. Similarly, a  quadripartition-based  heterogeneous Dice loss is:
\begin{equation}\label{eq:l_dice_l}
 \mathcal{L}_{dice}^{lqh}(p, y^*) = 1 - \frac{1}{C}\sum_{c=0}^{C-1}\frac{\sum_{\Omega_t\in\Omega_{all}}w_t\sum_{\bold{x} \in \Omega_t} 2 y^*_{\bold{x},c} \cdot p_{\bold{x},c} }{\sum_{\Omega_t\in\Omega_{all}}w_t\sum_{\bold{x} \in \Omega_t}  (y^*_{\bold{x},c} + p_{\bold{x},c})} 
\end{equation}

\subsection{Adaptive  Threshold for Confidence Estimation}
To identify confident regions in a prediction to obtain $\Omega_{UC}$, $\Omega_{US}$, $\Omega_{DC}$ and $\Omega_{DS}$, a threshold is usually needed for a pixel-wise confidence score such as the maximal probability or entropy. However, using a fixed confidence threshold may require manual tuning, which is not only time-consuming, but also insufficient to deal with the imbalance among multiple classes. For example, for easy classes, as most pixels will have highly confident predictions, a relatively higher confidence threshold is desired. For hard classes such as small targets, a relatively lower confidence threshold is expected to avoid the confident region being too small to obtain sufficient supervision signal for such classes. In addition, the overall confidence is usually small at the early training stage, and gradually increases at a later training stage. Therefore,  an adaptive threshold that changes with different classes and training stages is desirable for better identifying confident and suspicious regions of each class in our spatially heterogeneous loss. Despite that adaptive thresholds have previously been proposed by  FlexMatch~\cite{zhang2021flexmatch} and FreeMatch~\cite{wang2022freematch}, they were designed for image classification tasks. FlexMatch~\cite{zhang2021flexmatch} adjusts the threshold for a class by the ratio of sample number in that class. For segmentation tasks, the pixel number in different classes is highly imbalanced, which leads to extremely low threshold for minority classes, resulting in low quality of the filtered predictions. FreeMatch~\cite{wang2022freematch} defines the class-wise local threshold based on a relative prediction score among different classes, where the high degree of coupling of thresholds for different classes reduces the adaptability.  


Unlike these works, we propose a more effective adaptive thresholding strategy for $\gamma_{c^*}$ in segmentation tasks that is less sensitive to class imbalance and without interference between classes. 
Specifically, it is aware of the average confidence of each class and the training process simultaneously. Let $p_{\bold{x},c}$ denote the probability of class $c$ at pixel $\bold{x}$ in a segmentation probability prediction $p$, the corresponding label map by argmax is $y = \text{argmax}(p)$. The mean probability value for class $c$ in the prediction is:
\begin{equation}
O_c = \frac{\sum_{\bold{x}} (y_\bold{x} == c)\cdot p_{\bold{x},c}}{\sum_{\bold{x}} (y_\bold{x} == c)}
\end{equation}

Considering that $p$ and $O_c$ will change at different training iterations, we use $O_{c,i}$ to denote $O_{c}$ at the $i$-th iteration. Let $\gamma_{c,i}$ denote the confidence threshold of class $c$ at iteration $i$, it is updated based on an Exponential Moving Average (EMA):


\begin{equation}
\text{$\gamma$}_{c,i} = \alpha \cdot O_{c,i} + (1 - \alpha) \cdot \text{$\gamma$}_{c,i-1}
\end{equation}
where $\alpha \in $ [0, 1] denotes the  weight parameter for EMA, and the initial value of $\gamma_c$ is set as $\gamma_{c,0}=0.5$.

\begin{table*}[htbp]
\caption{Quantitative evaluation of different SSL methods on the ACDC dataset. SL: fully supervised learning from the labeled images. FullSup: fully supervised learning on all images. DSC: Dice Similarity Coefficient. HD$_{95}$: 95\% Hausdorff Distance.}
\scalebox{0.82}{
  \begin{tabular}{c|cccc|cccc|cccc|cccc}
    \toprule
    \multirow{3}[6]{*}{Method} & \multicolumn{8}{c|}{1 annotated volume} & \multicolumn{8}{c}{3 annotated volumes} \\
    \cmidrule{2-17} & \multicolumn{4}{c|}{DSC (\%) \(\uparrow\)} & \multicolumn{4}{c|}{HD$_{95}$ (mm) \(\downarrow\)} & \multicolumn{4}{c|}{DSC (\%) \(\uparrow\)} & \multicolumn{4}{c}{HD$_{95}$ (mm) \(\downarrow\)} \\
    \cmidrule{2-17} & \multicolumn{1}{c}{RV} & \multicolumn{1}{c}{MYO} & \multicolumn{1}{c}{LV} & Average & \multicolumn{1}{c}{RV} & \multicolumn{1}{c}{MYO} & \multicolumn{1}{c}{LV} & Average & \multicolumn{1}{c}{RV} & \multicolumn{1}{c}{MYO} & \multicolumn{1}{c}{LV} & Average & \multicolumn{1}{c}{RV} & \multicolumn{1}{c}{MYO} & \multicolumn{1}{c}{LV} & Average \\
    \midrule
   SL &  23.42 & 32.31 & 40.13 & 31.95$_{\pm25.31}$ & 72.08 & 53.22 & 53.13 & 59.47$_{\pm21.38}$ & 44.24 & 56.37 & 66.30 & 55.64$_{\pm25.95}$ & 52.84 & 13.61 & 24.53 & 30.33$_{\pm26.55}$\\
  UAMT~\cite{Yu2019miccai} & 21.32 & 32.24 & 29.16 & 27.58$_{\pm23.57}$ & 79.36 & 54.45 & 65.61 & 66.48$_{\pm25.20}$ & 46.16 & 67.66 & 78.49 & 64.10$_{\pm23.10}$ & 22.01 & 10.86 & 18.96 & 17.27$_{\pm25.32}$\\
   CCVC~\cite{wang2023conflict} & 19.58 & 37.79 & 41.29 & 32.89$_{\pm21.16}$ & 48.69 & 28.67 & 25.99 & 34.45$_{\pm35.27}$ & 58.62 & 71.35 & 79.82 & 69.93$_{\pm20.89}$ & 54.23 & 20.38 & 14.39 & 29.66$_{\pm28.81}$\\
   DCNet~\cite{chen2023decoupled} & 38.69 & 46.80 & 53.38 & 46.29$_{\pm31.99}$ & 39.33 & 22.82 & 30.65 & 30.93$_{\pm27.75}$ & 71.32 & 78.82 & 86.53 & 78.89$_{\pm15.28}$ & 12.43 & 5.32 & 8.41 & 8.72$_{\pm11.87}$\\
  { URPC~\cite{media2022urpc}} & {22.55} & {40.17} & {54.86} & {39.19}$_{{\pm}26.29}$ & {96.19} & {32.03} & {34.44} & {54.22}$_{{\pm}29.25}$ & {82.81} & {36.46} & {39.10} & {52.79}$_{{\pm}28.18}$ & {22.12} & {15.02} & {17.43} & {18.19}$_{{\pm}22.47}$\\

{CEUD~\cite{li2023confidence}} & {31.29} & {66.14} & {77.76} & {58.59}$_{{\pm}23.24}$ & {29.16} & {28.27} & {35.15} & {30.87}$_{{\pm}31.82}$ & {62.68} & {75.05} & {83.09} & {73.61}$_{{\pm}21.01}$ & {25.71} & {6.32} & {11.26} & {14.43}$_{{\pm}17.34}$\\

{SDCL~\cite{song2024sdcl}} & {65.53} & {65.79} & {79.85} & {70.40}$_{{\pm}21.22}$ & {12.30} & {\textbf{2.19}} & {3.12} & {5.87}$_{{\pm}14.42}$ & {75.99} & {75.41} & {88.16} & {79.86}$_{{\pm}13.24}$ & {6.12} & {2.30} & {\textbf{1.56}} & {3.32}$_{{\pm}5.93}$\\
   \midrule
   CCT~\cite{ouali2020semi} & 30.30 & 38.25 & 43.76 & 37.44$_{\pm27.84}$ & 70.00 & 36.16 & 38.16 & 48.11$_{\pm33.41}$ & 49.96 & 62.37 & 71.83 & 61.39$_{\pm25.71}$ & 28.05 & 8.54 & 17.62 & 18.07$_{\pm22.87}$\\
   +MetaSSL & 72.53 & 74.86 & 83.47 & 76.96$_{\pm14.01}$ & 7.71 & 13.25 & 12.25 & 11.07$_{\pm16.54}$ & 83.37 & 83.49 & 90.03 & 85.63$_{\pm8.37}$ & 2.24 & 3.04 & 6.59 & 3.95$_{\pm6.00}$\\
   \midrule
   MT~\cite{tarvainen2017mean} & 31.29 & 37.52 & 45.65 & 38.15$_{\pm23.95}$ & 75.46 & 59.07 & 63.82 & 66.12$_{\pm27.94}$ & 49.79 & 62.43 & 75.74 & 62.65$_{\pm22.55}$ & 41.48 & 20.54 & 35.57 & 32.53$_{\pm28.57}$\\
   +MetaSSL  & 45.79 & 78.67 & 87.59 & 71.68$_{\pm15.14}$ & 79.01 & 4.54 & 4.32 & 29.29$_{\pm16.56}$ & 85.72 & 84.30 & 90.09 & 86.70$_{\pm7.31}$ & 2.90 & 4.33 & 4.37 & 3.87$_{\pm7.40}$\\
   \midrule
   CPS~\cite{chen2021semi} & 23.14 & 39.79 & 40.50 & 34.47$_{\pm20.97}$ & 94.23 & 65.58 & 55.04 & 71.61$_{\pm17.87}$ & 57.45 & 59.75 & 75.64 & 64.28$_{\pm20.71}$ & 13.93 & 11.55 & 18.49 & 14.66$_{\pm20.42}$\\
   +MetaSSL  & \textbf{77.98} & \textbf{82.08} & \textbf{89.48} & \textbf{83.17$_{\pm9.69}$} & \textbf{4.00} & 2.67 & \textbf{1.66} & \textbf{2.70$_{\pm3.21}$} & \textbf{87.07} & 86.13 & 91.73 & \textbf{88.31$_{\pm5.79}$} & \textbf{1.65} & \textbf{1.23} & 2.11 & \textbf{1.66$_{\pm1.80}$}\\
    \midrule
    R-drop~\cite{liang2021r} & 34.33 & 41.20 & 48.20 & 41.24$_{\pm25.53}$ & 57.02 & 35.00 & 45.17 & 45.73$_{\pm31.67}$ & 53.12 & 63.12 & 76.51 & 64.25$_{\pm22.68}$ & 32.44 & 7.63 & 9.94 & 16.67$_{\pm19.22}$\\
    +MetaSSL& 76.59 & 78.65 & 88.57 & 81.27$_{\pm10.48}$ & 8.21 & 4.58 & 5.90 & 6.23$_{\pm12.28}$ & 84.64 & 83.91 & 91.03 & 86.53$_{\pm6.93}$ & 2.04 & 3.38 & 2.27 & 2.56$_{\pm4.55}$\\
   \midrule
   FixMatch~\cite{sohn2020fixmatch} & 19.83 & 58.06 & 68.23 & 48.71$_{\pm15.44}$ & 50.34 & 38.12 & 52.59 & 47.02$_{\pm31.00}$ & 52.17 & 71.78 & 78.88 & 67.61$_{\pm14.39}$ & 60.70 & 20.07 & 26.44 & 35.74$_{\pm25.26}$\\
    +MetaSSL& 75.24 & 76.19 & 85.21 & 78.88$_{\pm9.89}$ & 23.46 & 17.75 & 17.47 & 19.56$_{\pm18.64}$ & 74.93 & 78.34 & 84.67 & 79.31$_{\pm9.73}$ & 13.58 & 24.04 & 14.26 & 17.29$_{\pm16.76}$ \\
    \midrule
    BCP~\cite{bai2023bidirectional} & 67.56 & 66.75 & 59.42 & 64.58$_{\pm24.88}$ & 59.43 & 16.48 & 76.48 & 50.80$_{\pm30.42}$ & 85.51 & 85.73 & 90.86 & 87.36$_{\pm7.06}$ & 2.04 & 1.28 & 3.47 & 2.26$_{\pm2.98}$\\
    +MetaSSL & 76.28 & 79.95 & 74.48 & 76.90$_{\pm12.66}$ & 29.14 & 11.33 & 13.49 & 19.99$_{\pm17.99}$ & 86.29 & \textbf{86.53} & \textbf{91.83} & 88.22$_{\pm6.68}$ & 2.02 & 1.78 & 3.36 & 3.87$_{\pm7.40}$\\
   \midrule
   FullSup & 90.58 & 89.03 & 94.31 & 91.31$_{\pm4.33}$ & 1.26 & 1.59 & 2.24 & 1.70$_{\pm3.66}$ & 90.58 & 89.03 & 94.31 & 91.31$_{\pm4.33}$ & 1.26 & 1.59 & 2.24 & 1.70$_{\pm3.66}$\\
    \bottomrule
\end{tabular}
}
\label{tab:sota_ACDC}%
\end{table*}

\subsection{Applying MetaSSL to Existing SSL Frameworks}\label{sec:backbone_framework}
Based on the above quadripartition-based heterogeneous loss for labeled and unlabeled images, our MetaSSL can be applied to different existing multi-prediction-based SSL frameworks. The general loss for MetaSSL is defined as:
\begin{equation}
    \mathcal{L}=\mathcal{L}^{lqh}   + \lambda \mathcal{L}^{uqh}
\end{equation}
where $\lambda$ is a hyper-parameter to control the relative weight between unsupervised loss $\mathcal{L}^{uqh}$ defined on unlabeled images and supervised loss $\mathcal{L}^{lqh}$ defined on labeled images. 
Following the general practice of supervised learning~\cite{isensee2021nnu} that combines cross entropy loss and Dice loss for labeled images, the supervised loss in this work is defined as:
\begin{equation}
\mathcal{L}^{lqh}=\mathcal{L}^{lqh}_{ce}+ \mathcal{L}^{lqh}_{dice}
\end{equation}
where $\mathcal{L}^{lqh}_{ce}$ and $\mathcal{L}^{lqh}_{dice}$ are given in Eq.~\eqref{eq:l_ce_l} and Eq.~\eqref{eq:l_dice_l}, respectively. $\mathcal{L}^{uqh}$ is formulated in Eq.~\eqref{eq:L_unsup}, and it is defined between $p$ and $\tilde{p}$. Note that $p$ and $\tilde{p}$ can be obtained by different strategies according to the backbone SSL framework, as detailed in the following:

\textbf{MetaSSL + MT.}
For Mean Teacher (MT) framework~\cite{tarvainen2017mean}, as the teacher is an EMA of the student, the loss for back-propagation is only defined for the student only. We use $p$ and $\tilde{p}$ to denote the prediction from the student and teacher, respectively. 

\textbf{MetaSSL + FixMatch.} In FixMatch~\cite{sohn2020fixmatch}, the prediction for a weakly augmented image is used to supervise that for a strongly augmented version, and we denote the prediction for the weakly and strongly augmented versions as $p$ and $\tilde{p}$, respectively.  

\textbf{MetaSSL + CCT.}
As the Cross Consistency-Training (CCT)~\cite{ouali2020semi} method has a primary decoder and multiple auxiliary decoders, we use $\tilde{p}$ and $p$ to denote the prediction of the  primary decoder and a certain auxiliary decoder, respectively. The unsupervised loss $\mathcal{L}^{uqh}(p, \tilde{p})$ is applied to each auxiliary decoder. 

\textbf{MetaSSL + CPS.} The Cross Pseudo Supervision (CPS)~\cite{chen2021semi} method uses two networks to supervise each other on unlabeled images, and the predictions of the two networks are denoted as $p$ and $\tilde{p}$, respectively. The unsupervised loss is $\mathcal{L}^{uqh}(p, \tilde{p})$ + $\mathcal{L}^{uqh}(\tilde{p}, p)$ for the cross supervision. 

\textbf{MetaSSL + R-Drop.} The R-Drop~\cite{liang2021r} method minimizes the bidirectional Kullback-Leibler (KL) divergence between two probability predictions  from the same network under two forward passes with dropout.  We denote the two predictions as $p$ and $\tilde{p}$ respectively, and replace the bidirectional KL loss by $\mathcal{L}^{uqh}(p, \tilde{p})$ + $\mathcal{L}^{uqh}(\tilde{p}, p)$  in this work.

\textbf{MetaSSL + BCP.} The BCP method~\cite{bai2023bidirectional} is an extension of MT with copy-and-paste between labeled and unlabeled images, and pseudo-label loss is used between the teacher and student. As each mixed image has a labeled region and an unlabeled region, we use $\mathcal{L}^{lqh}$ in the labeled region, and apply $\mathcal{L}^{uqh}$
only in the unlabeled region.

\section{EXPERIMENTS AND RESULTS}
\subsection{Datasets and Implementation Details}
\subsubsection{ACDC Dataset}
The ACDC dataset~\cite{bernard2018deep} consists of 
200  Magnetic Resonance Imaging (MRI)  scans obtained from 100 patients for segmentation of three cardiac substructures: the Left Ventricle (LV), Right Ventricle (RV), and Myocardium (MYO). We randomly partitioned the dataset at the patient level into  70\% for training, 10\% for validation, and 20\% for testing, respectively. For SSL, we considered two different annotation ratios of the training set: 1 volume and 3 volumes, which corresponds to 32 and 68 labeled slices, respectively. Given the relatively large inter-slice spacing (5-10 mm), we employed UNet network~\cite{ronneberger2015u} for slice-wise segmentation, and the results of slices in a scan were stacked into a 3D volume for quantitative evaluation. For preprocessing, all slices were resized to a uniform size of 256$\times$256 pixels.

\subsubsection{DigestPath Dataset}
It is a pathological image dataset comprising Whole-Slide Images (WSI) for binary segmentation of tumor lesions in colonoscopy~\cite{da2022digestpath}. The WSIs originate from four medical institutions and were captured at a magnification of $\times$20, with a pixel resolution of 0.475µm. To segment tumor regions, we randomly divided a total of 130 malignant WSIs into 100 for training, 10 for validation, and 20 for testing, respectively. For SSL, we explored two annotation ratios: 5\% and 10\%. For computational convenience, we further crop the WSIs into patches of size 256 $\times$ 256, resulting in a training set comprising 9,627 patches. At the testing stage, we evaluated our model at the WSI level.

\subsubsection{LA Dataset}
{The Left Atrium (LA) segmentation challenge dataset~\cite{xiong2021global} consists of 100 gadolinium-enhanced 3D Magnetic Resonance Imaging (GE-MRI) scans with expert annotations. Following existing works~\cite{luo2021semi,Yu2019miccai}, we used 80 scans for training and 20 for testing. For pre-processing, each volume was cropped to focus on the cardiac region, and normalized by intensity mean and standard deviation.}

\begin{figure}
    \centering
    \includegraphics[width=1\linewidth]{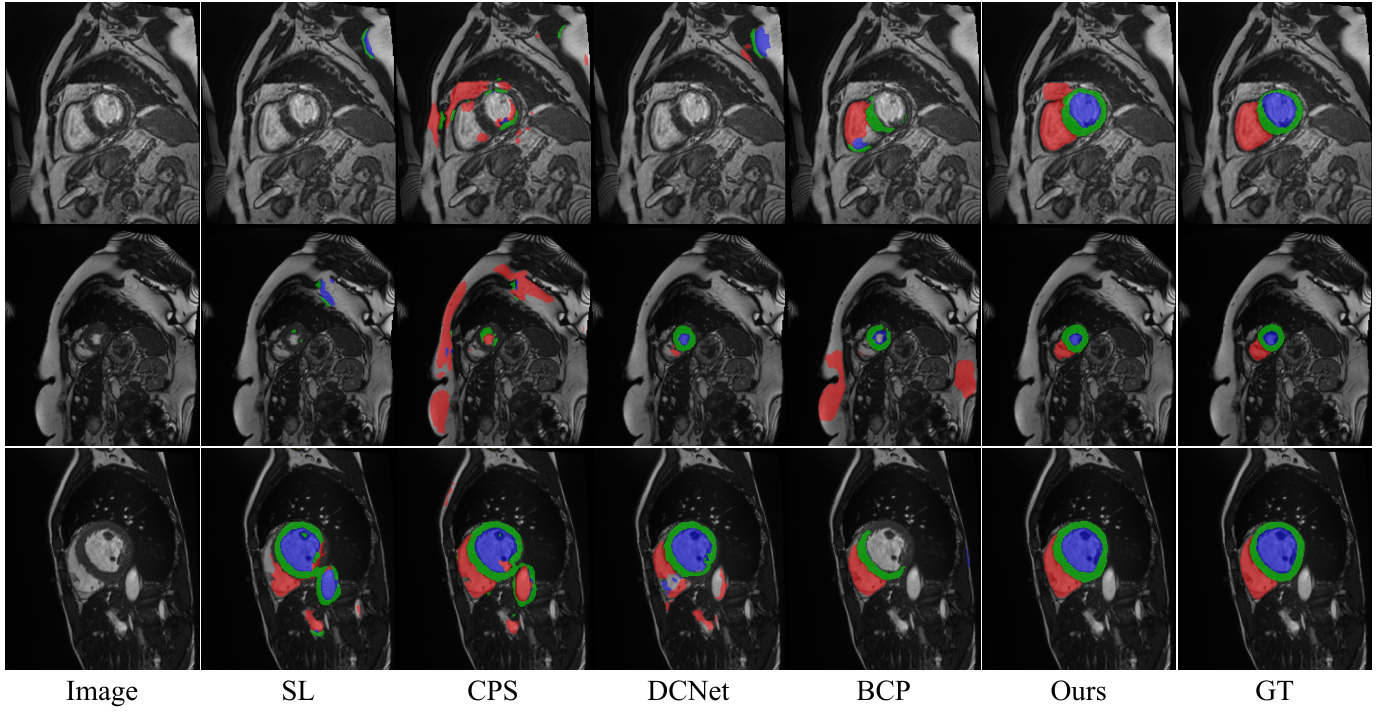}
    \caption{Visual comparison of different SSL methods on the ACDC dataset with one annotated volume. Ours refers to BCP + MetaSSL.}
    \label{fig:acdc}
\end{figure}

\begin{figure*}
    \centering
    \includegraphics[width=1\linewidth]{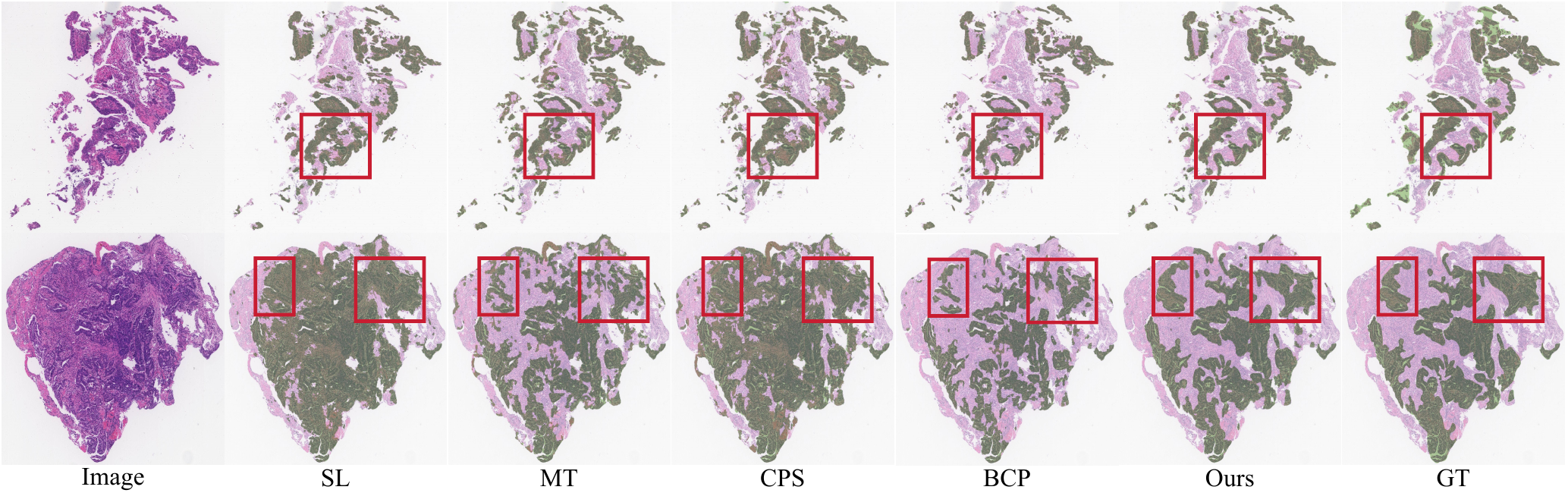}
    \caption{Visual comparison between different SSL methods on the DigestPath dataset with an annotation ratio of 5\%. Ours refers to BCP + MetaSSL. }
    \label{fig:path}
\end{figure*}

\subsubsection{Implementation Details}
 Our MetaSSL was implemented with different backbone SSL methods as described in Section~\ref{sec:backbone_framework} using the PyTorch framework and PyMIC\footnote{https://github.com/HiLab-git/PyMIC}~\cite{wang2022pymic} on an Ubuntu desktop equipped with a GTX2080TI GPU. Training was conducted by using the Stochastic Gradient Descent (SGD) optimizer, with a batch size of 24, where 12 slices/patches were labeled. The learning rate was initialized to 0.01 and adjusted based on a polynomial learning rate scheduler. The weighting factor $\lambda$ was defined by a time-dependent Gaussian warm-up function $\lambda(t) = 0.1 \cdot e^{-5(1-{t}/t_{max})^2}$, where $t$ represents the current training epoch, and $t_{max}$=600 was the total number of epochs. The value of \(\alpha\) was 0.99 following MT~\cite{tarvainen2017mean}, and {\(\beta\), \(\delta_u\), and \(\delta_l\) were set to 3, 0.3, and 0.6 based on optimal performance in the ablation study on the validation set.} After training, the model was deployed at the SenseCare~\cite{SenseCare2024} platform to support clinical research.
  For evaluation of the segmentation result, we used the Dice Similarity Coefficient (DSC) and the 95\% Hausdorff Distance (HD$_{95}$) for the ACDC dataset.  For the DigestPath dataset without a fixed segmentation structure, we employed the  DSC and the Jaccard Index (JI) following the practice in~\cite{zhong2023semi}.

\subsection{Comparison with Existing Semi-Supervised Methods}

In the experiments of SSL, we set the baseline as fully supervised learning from the labeled images in $\mathcal{D}_l$ only with a standard CE + Dice loss, which is referred as SL. {Additionally, we consider fully supervised learning on all images $\mathcal{D}$, termed FullSup.} We then compared it with 12 existing SSL methods:  {FixMatch~\cite{sohn2020fixmatch}, CCT~\cite{ouali2020semi}, MT~\cite{tarvainen2017mean}, CPS~\cite{chen2021semi}, UAMT~\cite{Yu2019miccai}, R-Drop~\cite{liang2021r},   URPC~\cite{media2022urpc}, CCVC~\cite{wang2023conflict}, DCNet~\cite{chen2023decoupled} and  BCP~\cite{bai2023bidirectional}, CEUD~\cite{li2023confidence} and SDCL~\cite{song2024sdcl}}. They were implemented with the same backbone network of UNet~\cite{ronneberger2015u} for fair comparison. 
{To validate the general applicability of our method to various multi-prediction-based SSL methods, we applied it to $6$ different SSL backbones, including  MT~\cite{tarvainen2017mean}, FixMatch~\cite{sohn2020fixmatch}, CCT~\cite{ouali2020semi}, R-Drop~\cite{liang2021r}, CPS~\cite{chen2021semi} and BCP~\cite{bai2023bidirectional}.}


\subsubsection{Results on ACDC Dataset}

\autoref{tab:sota_ACDC} shows quantitative results on the ACDC dataset. With one annotated volume, the baseline method SL obtained a very low performance with average DSC of 31.95\%. 
Most existing SSL methods improved it to some degree. The average DSC of FixMatch, CCT, MT and CPS ranged  from 34.47\% to 48.71\%.  {
Our proposed MetaSSL framework consistently demonstrates significant performance improvements across various SSL frameworks. When integrated with the CCT and MT frameworks, the DSC was significantly improved from 37.44\% and 38.15\% to 76.96\% and 71.68\%, respectively. In the CPS~\cite{chen2021semi} framework,  MetaSSL  improved the DSC by 48.70 percentage points, i.e., from 34.47\% to 83.17\%.  MetaSSL + CPS also significantly outperformed the best existing method SDCL~\cite{song2024sdcl} that obtained an average DSC of 70.40\%. Under a three-volume annotation budget, MetaSSL also delivers consistent performance gains across different existing SSL frameworks. Note that the MetaSSL-enhanced CPS framework achieved an average DSC of 88.31\%, and largely outperformed SDCL (79.86\%).  }
\autoref{tab:sota_ACDC} also shows that the $HD_{95}$ obtained by our method is much lower than that of the  corresponding backbone SSL method, especially when the annotation budget was one volume. The visual comparison in \autoref{fig:acdc} shows that the existing methods obtained severe under-segmentation or over-segmentation when the annotation budget is low, while our method could obtain more robust results.

\begin{table}
  \centering
  \caption{Quantitative comparison of different SSL methods on the DigestPath Dataset. {* indicates a significant improvement (p-value < 0.05) compared to the original method without MetaSSL using paired Student's t-test. SL: fully supervised learning from the labeled images. FullSup: fully supervised learning on all images. DSC: Dice Similarity Coefficient. HD$_{95}$: 95\% Hausdorff Distance.}}
    \scalebox{0.88}{\begin{tabular}{c|cc|cc}
    \toprule
\multirow{2}[6]{*}{Methods}&  \multicolumn{2}{c|}{DSC (\%) \(\uparrow\)} & \multicolumn{2}{c}{Jaccard Index (\%) \(\uparrow\)} \\
\cmidrule{2-5}          &  \multicolumn{1}{c|}{5\% labeled} & 10\% labeled & \multicolumn{1}{c|}{5\% labeled} & 10\% labeled \\
    \midrule
   SL & \multicolumn{1}{c|}{60.73±22.12} &  67.03±18.06     & \multicolumn{1}{c|}{47.06±22.01} &52.92±18.72  \\
       UAMT~\cite{Yu2019miccai}     & \multicolumn{1}{c|}{63.44±20.37} & 67.26±17.43  & \multicolumn{1}{c|}{49.47±20.49} & 53.08±18.60 \\
      CCVC~\cite{wang2023conflict}     &  \multicolumn{1}{c|}{67.32±18.09} &  68.66±16.39  & \multicolumn{1}{c|}{53.26±18.75} & 54.46±17.68\\
      DCNet~\cite{chen2023decoupled}      & \multicolumn{1}{c|}{67.80±17.38} &  69.75±16.74  & \multicolumn{1}{c|}{53.58±17.65} & 55.84±18.00 \\
     { URPC~\cite{media2022urpc}} & \multicolumn{1}{c|}{{66.53±18.17}} &  {68.29±18.30}  & \multicolumn{1}{c|}{{52.38±18.74}} & {54.45±19.03} \\
       {CEUD~\cite{li2023confidence}}    & \multicolumn{1}{c|}{{68.24±16.93}} &  {69.87±16.72}  & \multicolumn{1}{c|}{{54.34±17.90}} & {55.74±17.88} \\
     {SDCL~\cite{song2024sdcl}}    & \multicolumn{1}{c|}{{69.95±15.98}} &  {70.83±15.73}  & \multicolumn{1}{c|}{{55.83±16.46}} & {56.95±17.05} \\
       \midrule
       CCT~\cite{ouali2020semi}    & \multicolumn{1}{c|}{63.58±19.04} &  67.46±17.32     & \multicolumn{1}{c|}{49.30±19.41} & 53.28±18.40 \\
       +MetaSSL & \multicolumn{1}{c|}{69.46±15.94*} & 70.63±16.17* & \multicolumn{1}{c|}{55.27±17.08*} &  56.77±17.58*\\
       \midrule
       MT~\cite{tarvainen2017mean}    & \multicolumn{1}{c|}{65.05±19.13} & 67.66±16.98  & \multicolumn{1}{c|}{50.94±19.56} & 53.44±18.20 \\
       +MetaSSL  & \multicolumn{1}{c|}{68.55±18.14*} & 71.12±15.65* & \multicolumn{1}{c|}{54.73±18.93*} & 57.24±17.11* \\
       \midrule
       CPS~\cite{chen2021semi}    & \multicolumn{1}{c|}{65.32±18.05} &   68.15±17.15    & \multicolumn{1}{c|}{51.00±18.80} & 54.04±18.27 \\
       +MetaSSL   & \multicolumn{1}{c|}{70.02±16.63*} & 71.34±15.39* & \multicolumn{1}{c|}{56.12±17.79*} & 57.45±16.94* \\
       \midrule
        R-Drop~\cite{liang2021r}         & \multicolumn{1}{c|}{64.07±19.72} &  67.60±17.20  & \multicolumn{1}{c|}{50.00±19.98} & 53.40±18.25 \\
       +MetaSSL  & \multicolumn{1}{c|}{69.63±17.13*} & 71.10±15.87* & \multicolumn{1}{c|}{55.75±18.02*} & 57.26±17.24* \\
       \midrule
       FixMatch~\cite{sohn2020fixmatch}    & \multicolumn{1}{c|}{65.73±18.38} &   68.24±17.28   & \multicolumn{1}{c|}{51.72±19.23} & 54.16±18.32 \\
       +MetaSSL  & \multicolumn{1}{c|}{69.13±17.43*} & 71.49±15.68* & \multicolumn{1}{c|}{55.23±18.24*} & 57.67±16.98* \\
       \midrule
       BCP~\cite{bai2023bidirectional}    & \multicolumn{1}{c|}{69.80±14.89} &  70.99±15.32  & \multicolumn{1}{c|}{55.44±16.19} & 56.98±16.67 \\
       +MetaSSL  & \multicolumn{1}{c|}{\textbf{70.44±15.31}} & \textbf{71.96±14.73} & \multicolumn{1}{c|}{\textbf{56.31±16.58}} &  \textbf{58.04±16.15}\\
    \midrule
    FullSup & \multicolumn{2}{c|}{74.31±13.24} & \multicolumn{2}{c}{60.71±15.30} \\
    \bottomrule
    \end{tabular}}%
  \label{tab:sota_path}%
\end{table}%

\subsubsection{Results on DigestPath Dataset}

\autoref{tab:sota_path} shows the quantitative evaluation of different SSL methods with the annotation ratio being 5\% and 10\%.  {When the annotation ratio was 5\%, the SL baseline achieved an average DSC score of 60.73\%. Most existing SSL methods improved the performance. In addition, integrating our MetaSSL framework with various existing SSL methods consistently improves performance.} {Specifically, when combined with CCT, MetaSSL boosts the DSC from 63.58\% to 69.46\%, marking the highest absolute improvement of 5.88\%. Furthermore, applying MetaSSL to BCP enhances the DSC to 70.44\%, surpassing the best existing SSL method SDCL (69.95\%). When increasing the annotation ratio to 10\%, MetaSSL continues to enhance performance across all evaluated SSL frameworks. Notably, R-Drop with MetaSSL achieves improvements of 3.50\% in DSC and 3.86\% in Jaccard Index. Additionally,  MetaSSL further improves the DSC of the current best method BCP from 70.99\% to 71.96\%, showing its generalizability to different SSL frameworks. For a more intuitive interpretation of the improvement brought by MetaSSL, we visualized the corresponding DSC values of \autoref{tab:sota_ACDC}  and \autoref{tab:sota_path} in \autoref{fig:different_backbone}.}

\autoref{fig:path} presents a visual comparison of these methods on two cases. It shows that the SL baseline and CPS tend to obtain some over-segmentation. BCP was better than the other existing methods, but led to some under-segmentation. In contrast, the result of BCP + MetaSSL is closer to the ground truth with less mis-segmentation  of the tumor region. 

\begin{table}[tbp]
  \centering
  \caption{{Quantitative comparison of different SSL methods on the LA Dataset. SL: fully supervised learning from the labeled images. FullSup: fully supervised learning on all images. DSC: Dice Similarity Coefficient. HD$_{95}$: 95\% Hausdorff Distance.}}
   \scalebox{0.95}{
   \begin{tabular}{c|cc|cc}
    \toprule
\multirow{2}[3]{*}{{Methods}}&  \multicolumn{2}{c|}{{DSC (\%) \(\uparrow\)}} & \multicolumn{2}{c}{{HD$_{95}$ (mm) \(\downarrow\)}} \\
\cmidrule{2-5}          &  \multicolumn{1}{c|}{{2 volumes}} & {4 volumes} & \multicolumn{1}{c|}{{2 volumes}} & {4 volumes} \\
    \midrule
  {SL} & \multicolumn{1}{c|}{{59.03±16.83}} & {68.21±13.74} & \multicolumn{1}{c|}{{12.99±17.02}} & {11.06±15.78}\\
{URPC~\cite{media2022urpc}} & \multicolumn{1}{c|}{{74.41±11.20}} & {78.52±9.93} & \multicolumn{1}{c|}{{11.56±15.48}} & {10.89±9.98} \\
{SDCL~\cite{song2024sdcl}} & \multicolumn{1}{c|}{{85.50±7.08}} & {87.67±6.10} & \multicolumn{1}{c|}{{10.30±7.08}} & {9.47±6.15} \\
\midrule
{MT~\cite{tarvainen2017mean}} & \multicolumn{1}{c|}{{73.47±11.71}} & {77.62±11.85} & \multicolumn{1}{c|}{{12.49±18.53}} & {11.56±7.24} \\
{+MetaSSL } & \multicolumn{1}{c|}{{81.67±4.55}} & {86.32±4.38} & \multicolumn{1}{c|}{{10.73±11.58}} & {9.36±4.04} \\
\midrule
{CPS~\cite{chen2021semi}} & \multicolumn{1}{c|}{{77.12±11.53}} & {79.37±9.68} & \multicolumn{1}{c|}{{9.83±18.54}} & {9.26±14.59} \\
{+MetaSSL} & \multicolumn{1}{c|}{{82.17±8.57}} & {85.89±5.69} & \multicolumn{1}{c|}{{\textbf{9.33±16.01}}} & {8.60±8.18} \\
\midrule
{BCP~\cite{bai2023bidirectional}} & \multicolumn{1}{c|}{{84.70±5.19}} & {86.93±6.64} & \multicolumn{1}{c|}{{10.89±4.77}} & {10.39±4.23} \\
{+MetaSSL} & \multicolumn{1}{c|}{{\textbf{85.64±5.86}}} & {\textbf{87.82±6.24}} & \multicolumn{1}{c|}{{10.50±5.53}} & {\textbf{8.01±3.93}}\\

\midrule
{FullSup} & \multicolumn{2}{c|}{{90.32±4.89}} & \multicolumn{2}{c}{{5.91±3.25}} \\
    \bottomrule
    \end{tabular}
}%
  \label{tab:sota_LA}%
\end{table}%

\begin{figure}
    \centering
    \includegraphics[width=1\linewidth]{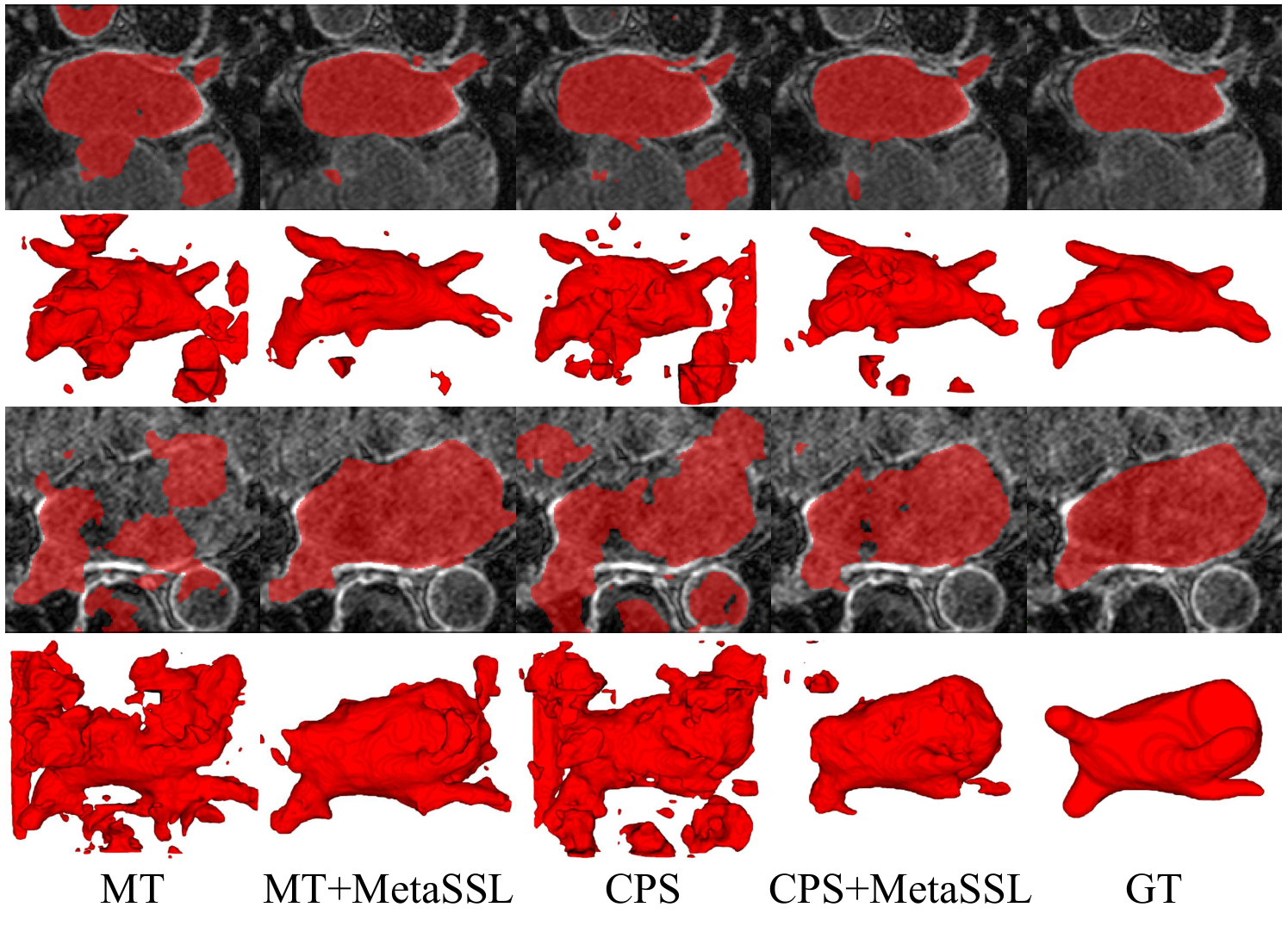}
    \caption{{Visual comparison of different SSL methods on the LA dataset with two annotated volumes.}}
    \label{fig:LA}
\end{figure}

\subsubsection{Results on LA Dataset}
{\autoref{tab:sota_LA} presents a quantitative comparison of different SSL methods on the LA dataset. The} {SL baseline achieved DSC scores of 59.03\% and 68.21\% under annotation budget of 2 and 4 volumes, respectively. While existing SSL methods demonstrated improvements over the baseline, our proposed MetaSSL framework consistently brings performance improvement to these SSL approaches. For MT~\cite{tarvainen2017mean}, MetaSSL significantly improves  the DSC from 73.47\% to 81.67\%, and from 77.62\% to 86.32\% under the two annotation budgets, respectively. Similarly, MetaSSL-augmented BCP achieved the highest average DSC scores of 85.64\% (2 annotated volume) and 87.82\% (4 annotated volume), surpassing both the original BCP and the other existing SSL methods. In terms of HD95 values, our MetaSSL also helps to improve the boundary accuracy when combined with different existing SSL methods. 
The qualitative comparison presented in \autoref{fig:LA} illustrates that both MT and CPS methods suffer from severe over-segmentation issues under limited annotation budgets, whereas MateSSL demonstrates more robust segmentation outcomes when applied to them.}

\begin{table*}[htbp]
\caption{Ablation study of our method on the ACDC Dataset with CPS as the backbone SSL method.}
\scalebox{0.78}{
  \begin{tabular}{c|cccc|cccc|cccc|cccc}
    \toprule
    \multirow{3}{*}{Method} & \multicolumn{8}{c|}{1 annotated volume} & \multicolumn{8}{c}{3 annotated volumes} \\
    \cmidrule{2-17} & \multicolumn{4}{c|}{DSC (\%) \(\uparrow\)} & \multicolumn{4}{c|}{HD$_{95}$ (mm) \(\downarrow\)} & \multicolumn{4}{c|}{DSC (\%) \(\uparrow\)} & \multicolumn{4}{c}{HD$_{95}$ (mm) \(\downarrow\)} \\
    \cmidrule{2-17} & RV & MYO & LV & Average &RV & MYO & LV & Average & RV & MYO & LV & Average & RV &MYO & LV & Average \\
    \midrule
    SL & 23.42 & 32.31 & 40.13 & 31.95$_{\pm25.31}$ & 72.08 & 53.22 & 53.13 & 59.47$_{\pm21.38}$ & 44.24 & 56.37 & 66.30 & 55.64$_{\pm25.95}$ & 52.84 & 13.61 & 24.53 & 30.33$_{\pm26.55}$ \\
    $\mathcal{L}^{lqh}$ & 47.33 & 64.20 & 78.65 & 63.39$_{\pm17.25}$ & 72.06 & 44.48 & 30.37 & 48.97$_{\pm31.42}$ & 66.28 & 77.71 & 87.60 & 76.42$_{\pm12.81}$ & 28.95 & 11.80 & 19.26 & 20.00$_{\pm24.11}$ \\
    \midrule
    CPS & 23.14 & 39.79 & 40.50 & 34.47$_{\pm20.97}$ & 94.23 & 65.58 & 55.04 & 71.61$_{\pm17.87}$ & 57.45 & 59.75 & 75.64 & 64.28$_{\pm20.71}$ & 13.93 & 11.55 & 18.49 & 14.66$_{\pm20.42}$ \\
    + $\mathcal{L}^{uqh}$ & 66.48 & 71.65 & 81.43 & 73.19$_{\pm22.31}$ & 8.93 & 10.29 & 8.00 & 9.07$_{\pm15.76}$ & 81.06 & 82.06 & 91.15 & 84.76$_{\pm8.92}$ & 2.55 & 3.35 & 7.19 & 4.36$_{\pm7.80}$ \\
    + $\mathcal{L}^{lqh} + \mathcal{L}^{uqh}$ & \textbf{77.98} & \textbf{82.08} & \textbf{89.48} & \textbf{83.17$_{\pm9.69}$} & \textbf{4.00} & \textbf{2.67} & \textbf{1.66} & \textbf{2.77$_{\pm3.21}$} & \textbf{87.07} & \textbf{86.13} & \textbf{91.73} & \textbf{88.31$_{\pm5.79}$} & \textbf{1.65} & \textbf{1.23} & 2.11 & \textbf{1.66$_{\pm1.80}$} \\
    \midrule
    + Ours (conf-bh) & 12.55 & 52.11 & 67.90 & 44.19$_{\pm21.91}$ & 95.42 & 35.19 & 45.84 & 58.82$_{\pm26.93}$ & 36.61 & 65.39 & 81.00 & 69.36$_{\pm12.50}$ & 87.38 & 5.20 & 4.10 & 32.23$_{\pm18.02}$ \\
    + Ours (cons-bh) & 36.88 & 82.11 & 89.08 & 56.70$_{\pm22.88}$ & 98.50 & 8.08 & 9.46 & 38.68$_{\pm17.61}$ & 54.46 & 83.36 & 90.05 & 75.94$_{\pm11.77}$ & 68.94 & 1.45 & \textbf{1.99} & 24.13$_{\pm15.87}$ \\
     \midrule
  \end{tabular}
}
\label{tab:Ablation}
\end{table*}

\subsection{Ablation Study}
In order to verify the effectiveness of each component in our method, we conducted ablation experiments using the backbone SSL method of CPS~\cite{chen2021semi}.

\subsubsection{Effectiveness our spatially heterogeneous loss on unlabeled and labeled images}
To validate the effectiveness of our method that applies the spatially heterogeneous loss for both labeled and unlabeled images, we compare it with two variants: 1) The original CPS method that uses spatially uniform loss for both labeled and unlabeled images; 2) CPS + $\mathcal{L}^{uqh}$ that applies our quadripartition-based heterogeneous loss only for unlabeled images. Correspondingly, our method is denoted as CPS + $\mathcal{L}^{uqh}$ + $\mathcal{L}^{lqh}$. The performance comparison of these methods on the ACDC dataset is shown in \autoref{tab:Ablation}.
When the annotation budget was one volume, the baseline CPS method obtained an average DSC score of 34.47\%. By replacing the original spatially uniform pseudo-label loss by our $\mathcal{L}^{uqh}$, the average DSC improved to 73.19\%. Additionally, further replacing the original spatially uniform supervised loss by our $\mathcal{L}^{lqh}$, the average DSC improved to 83.17\%. When the annotation budget was three volumes, the average DSC of original CPS was 64.28\%, and  CPS + $\mathcal{L}^{uqh}$ improved it to 84.76\%. Finally, CPS + $\mathcal{L}^{uqh}$ + $\mathcal{L}^{lqh}$ improved it to 88.31\%, showing the effectiveness of our spatially heterogeneous loss for labeled and unlabeled images. 

In addition, when only using the annotated images for training, we replaced the SL baseline by SL + $\mathcal{L}^{lqh}$, i.e., our spatially heterogeneous loss is used for supervised learning. \autoref{tab:Ablation} shows that $\mathcal{L}^{lqh}$ led to an improvement of 31.44 and 20.78 percentage points for SL in terms of DSC under the two annotation budgets, respectively.  

\begin{figure}
    \centering
    \includegraphics[width=1\linewidth]{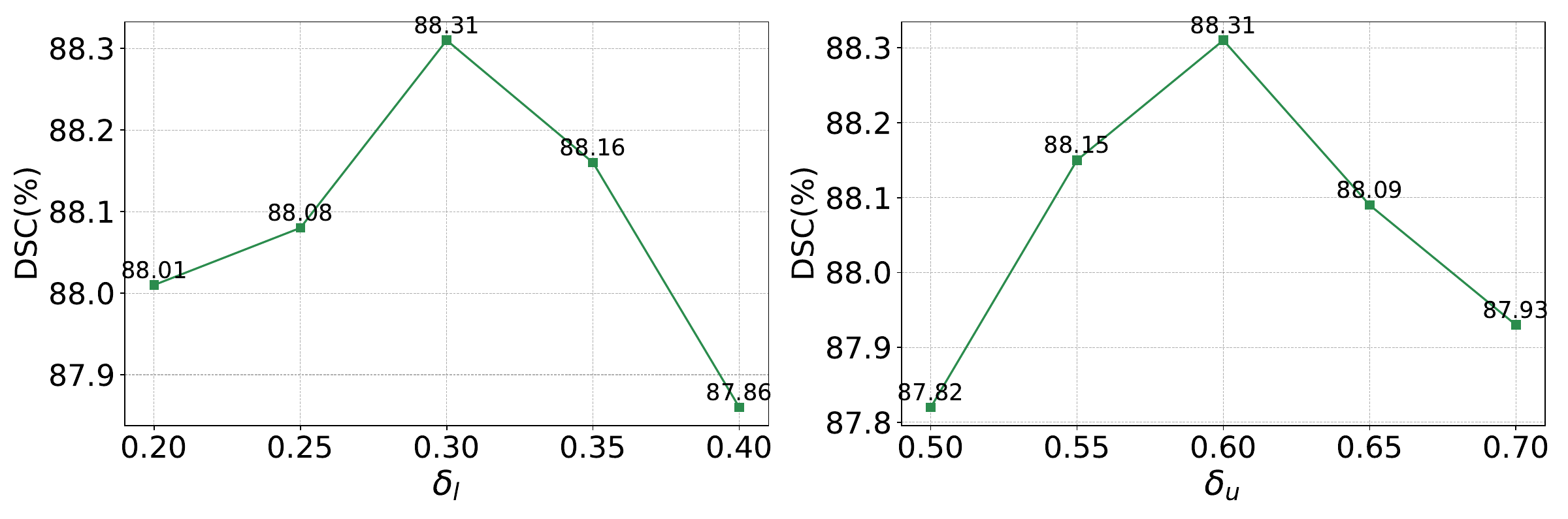}
    \caption{Effect of hyper-parameters $\delta_l$ and $\delta_u$ on the ACDC dataset with 3 annotated volumes.}
    \label{fig:parameters}
\end{figure}

\begin{table}[htbp]
  \centering
  \caption{Comparison between different methods for setting the confidence threshold on the ACDC dataset.}
    \scalebox{0.88}{\begin{tabular}{c|cc|cc}
    \toprule
    \multirow{2}[4]{*}{Methods}  &  \multicolumn{2}{c|}{1 annotated volume} & \multicolumn{2}{c}{3 annotated volumes} \\
\cmidrule{2-5}          &  \multicolumn{1}{c|}{DSC(\%) \(\uparrow\)} & HD$_{95}$(mm) \(\downarrow\) & \multicolumn{1}{c|}{DSC(\%) \(\uparrow\)} & HD$_{95}$(mm) \(\downarrow\) \\
    \midrule
   $\gamma_{c^*}$ = 0.85 & \multicolumn{1}{c|}{36.09±20.88} & 82.55±13.07 & \multicolumn{1}{c|}{77.29±13.03} & 16.95±19.44  \\
  $\gamma_{c^*}$ = 0.90    & \multicolumn{1}{c|}{43.00±23.20} & 50.57±24.61 & \multicolumn{1}{c|}{86.68±9.27} & 2.99±5.41 \\
     $\gamma_{c^*}$ = 0.95    & \multicolumn{1}{c|}{32.88±19.10} & 52.34±27.00 & \multicolumn{1}{c|}{85.90±10.06} & 3.40±6.11 \\
     \midrule
     {FlexMatch~\cite{zhang2021flexmatch}}  & \multicolumn{1}{c|}{{78.76±15.10}} & {4.87±6.95} & \multicolumn{1}{c|}{{83.93±11.12}} & {4.29±6.58} \\
    {FreeMatch~\cite{wang2022freematch}} & \multicolumn{1}{c|}{{81.04±10.30}} & {4.00±6.37} & \multicolumn{1}{c|}{{85.81±+9.91}} & {3.27±5.12} \\
      \midrule
    Adaptive $\gamma_{c^*}$ & \multicolumn{1}{c|}{\textbf{83.17±9.69}} & \textbf{2.77±3.21} & \multicolumn{1}{c|}{\textbf{88.31±5.79}} & \textbf{1.66±1.80} \\
    \bottomrule
    \end{tabular}}%
  \label{tab:rehold}%
\end{table}%

\begin{figure*}
    \centering
    \includegraphics[width=1\linewidth]{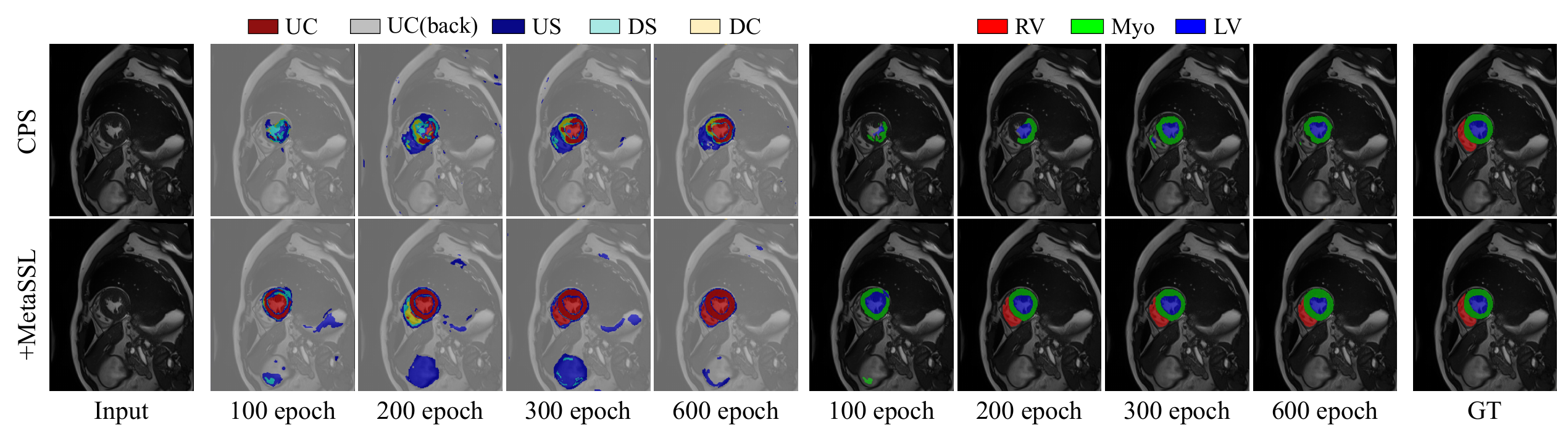}
    \caption{{Visualization of segmentation results and the four partitions (UC, US, DC, and DS) under the CPS framework with and without MetaSSL on the ACDC dataset across different training epochs.}}
    \label{fig:view1_new}
\end{figure*}

\subsubsection{The role of uncertainty and confidence} 
Our spatially heterogeneous losses $\mathcal{L}^{uqh}$ and  $\mathcal{L}^{lqh}$ rely on a quadripartition of the reference prediction based on confidence and consistency. In order to assess the role of each of them, we compared our method with two variants that only partitions the reference prediction into two regions: 1) Ours (conf-bh) that uses a bisection of the reference prediction based on a confidence threshold $\gamma_{c^*}$ to obtain confident and uncertain regions,  with confident region having a higher weight than the uncertain region in the spatially heterogeneous loss; 2) Ours (cons-bh) where a consistency-based bisection of the reference prediction is used to obtain unanimous and discrepant  regions for the spatially heterogeneous loss, with the former having a higher weight than the later. The results of these two variants are shown in the last section of \autoref{tab:Ablation}. It can be observed for both annotation budgets, Ours (conf-bh) and (cons-bh) outperformed the CPS baseline. In contrast, our quadripartition-based heterogeneous outperformed both of them, showing the superiority of combining confidence and consistency for quadripartition than only using one of them for bisection in the heterogeneous loss. 

\subsubsection{Effectiveness of adaptive confidence threshold}
To investigate the effectiveness of our class-wise adaptive confidence threshold $\gamma_{c^*}$, we compared it with a fixed threshold value by setting $\gamma_{c^*}$ to 0.85, 0.9 and 0.95, respectively. {In addition, we also used the adaptive confidence threshold strategies in FlexMatch~\cite{zhang2021flexmatch} and FreeMatch~\cite{wang2022freematch} for comparison.} 
As shown in \autoref{tab:rehold}, when the annotation budget was one volume, a fixed $\gamma_{c^*}$ led to quite low DSC scores (ranging from 32.88\% to 43.00\%). In contrast, {our adaptive $\gamma_{c^*}$ significantly improved the performance, with a DSC of 83.18\%, and it also outperformed FlexMatch~\cite{zhang2021flexmatch} and FreeMatch~\cite{wang2022freematch}.} When the annotation budget was three volumes,  a fixed $\gamma_{c^*}$  could lead to relatively high DSC scores (ranging from 77.29\% to 86.68\%), but the result is very sensitive to the value of $\gamma_{c^*}$.  In contrast, our adaptive $\gamma_{c^*}$  performed better than the fixed values, and it avoids manual tuning of $\gamma_{c^*}$.  
\begin{table}
  \centering
  \caption{Effect of function $\phi(u)$ on the ACDC dataset with 3 annotated volumes.}
    \scalebox{1}{\begin{tabular}{c|cc}
    \toprule
    Function        &  \multicolumn{1}{c|}{DSC(\%) \(\uparrow\)} & HD$_{95}$(mm) \(\downarrow\) \\
    \midrule
   Linear & \multicolumn{1}{c|}{69.76±17.30} &  11.16±15.64      \\
   {Reciprocal} & \multicolumn{1}{c|}{{74.07±17.41}} &  {18.07±15.97}      \\
   {Cosine} & \multicolumn{1}{c|}{{77.02±15.01}} &  {12.69±15.40}      \\
   Generalized Gaussian ($\beta$=1) & \multicolumn{1}{c|}{56.67±20.94} &   35.34±16.34  \\
  Standard Gaussian ($\beta$=2)  & \multicolumn{1}{c|}{79.81±10.49} &   11.04±19.87   \\
   Generalized Gaussian ($\beta$=3) & \multicolumn{1}{c|}{\textbf{84.13±10.97}} &   \textbf{5.52±7.70}   \\
    \bottomrule
    \end{tabular}}%
  \label{tab:down}%
\end{table}%

\subsubsection{Analysis of hyper-parameters}
Our method has three main hyper-parameters for  setting the weight of different regions: $\beta$ in the generalized Gaussian function,  and the intervals $\delta_l$ and $\delta_u$ for labeled and unlabeled images. {First, for the selection of the monotonically decreasing function $\phi(u)$ where the output value is in the range of (0, 1], we compared five variants: 1) A linearly decreasing function from (0.0, 1.0) to (2.0, 0.0); 2) The reciprocal function $\phi(u)=\left(u+1\right)^{-1}$; 3) The cosine function $\phi(u)=0.5\cdot\cos u+0.5$; 4) A standard Gaussian function with $\beta$=2, i.e., $\phi(u)=e^{-u^2}$; 5) A generalized Gaussian function with $\beta$=1, i.e., $\phi(u)=e^{-u}$; and 6) a generalized Gaussian function with $\beta$=3, i.e., $\phi(u)=e^{-u^3}$.} 
A comparison of these methods on the ACDC dataset when $\delta_l=\delta_u=0.6$ is shown in \autoref{tab:down}. It can be seen that for the generalized Gaussian function, $\beta$=3 obtained the best performance, and it also outperformed the linear function, reciprocal function and cosine function. 

Then we set $\beta$=3 and tuned $\delta_l$ and $\delta_u$ in the range of [0.2, 0.7]. The results in \autoref{fig:parameters} indicate that the optimal value for \( \delta_l \) and \( \delta_u \) was 0.3 and 0.6, respectively. 


\begin{figure}
    \centering
    \includegraphics[width=1\linewidth]{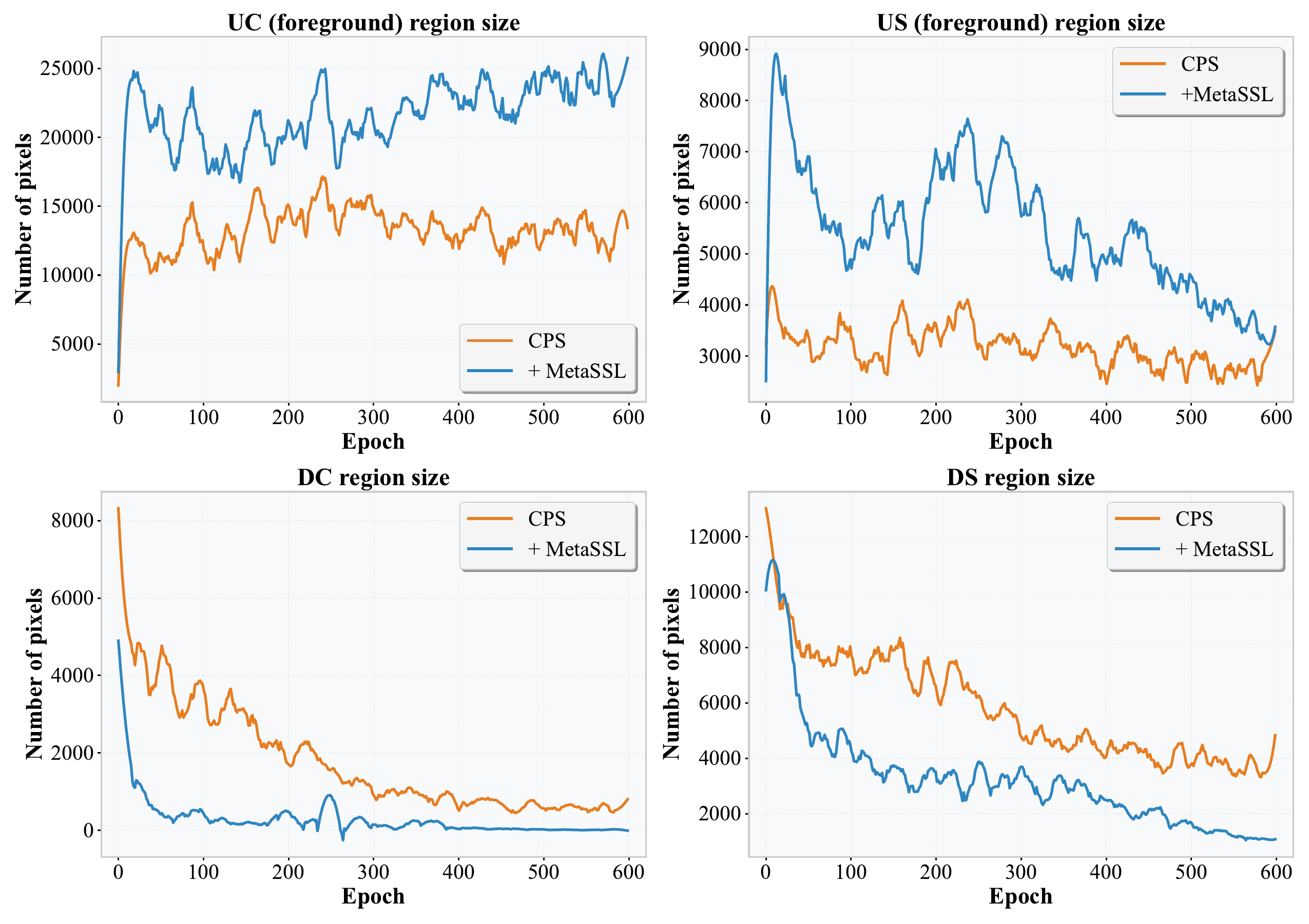}
    \caption{{The dynamic changes in pixel quantities within the UC, US, DC, and DS regions across different epochs in the ACDC dataset with three annotated volumes.}}
    \label{fig:Pixel_number}
\end{figure}

\subsubsection{Evolution of the four regions during training} { \autoref{fig:view1_new} shows a visualization of the quadripartition maps at different training epochs. It can be observed that the UC region increases during training for MetaSSL + CPS. Compared with the backbone SSL method CPS, the introduction of MetaSSL leads to better pseudo-labels with larger UC regions. 
\autoref{fig:Pixel_number} further shows the number of pixels in each of the UC, US, DC, and DS regions during training. Since the UC and US regions predominantly consist of background pixels, which could obscure meaningful comparisons, we restricted our analysis to foreground pixels for these two regions. It can be observed that the UC and US regions gradually increase and decrease respectively during training after using MetaSSL, while the original CPS has a} {relatively stable size of US and US regions, showing that MetaSSL is more effective in obtaining better results indicated by the increased consistency between models and the confidence. In addition, a decrease of DC and DS regions can be observed for both CPS and CPS + MetaSSL, and the later has a higher convergence speed, which helps better training.    } 

\begin{figure}
    \centering
    \includegraphics[width=1\linewidth]{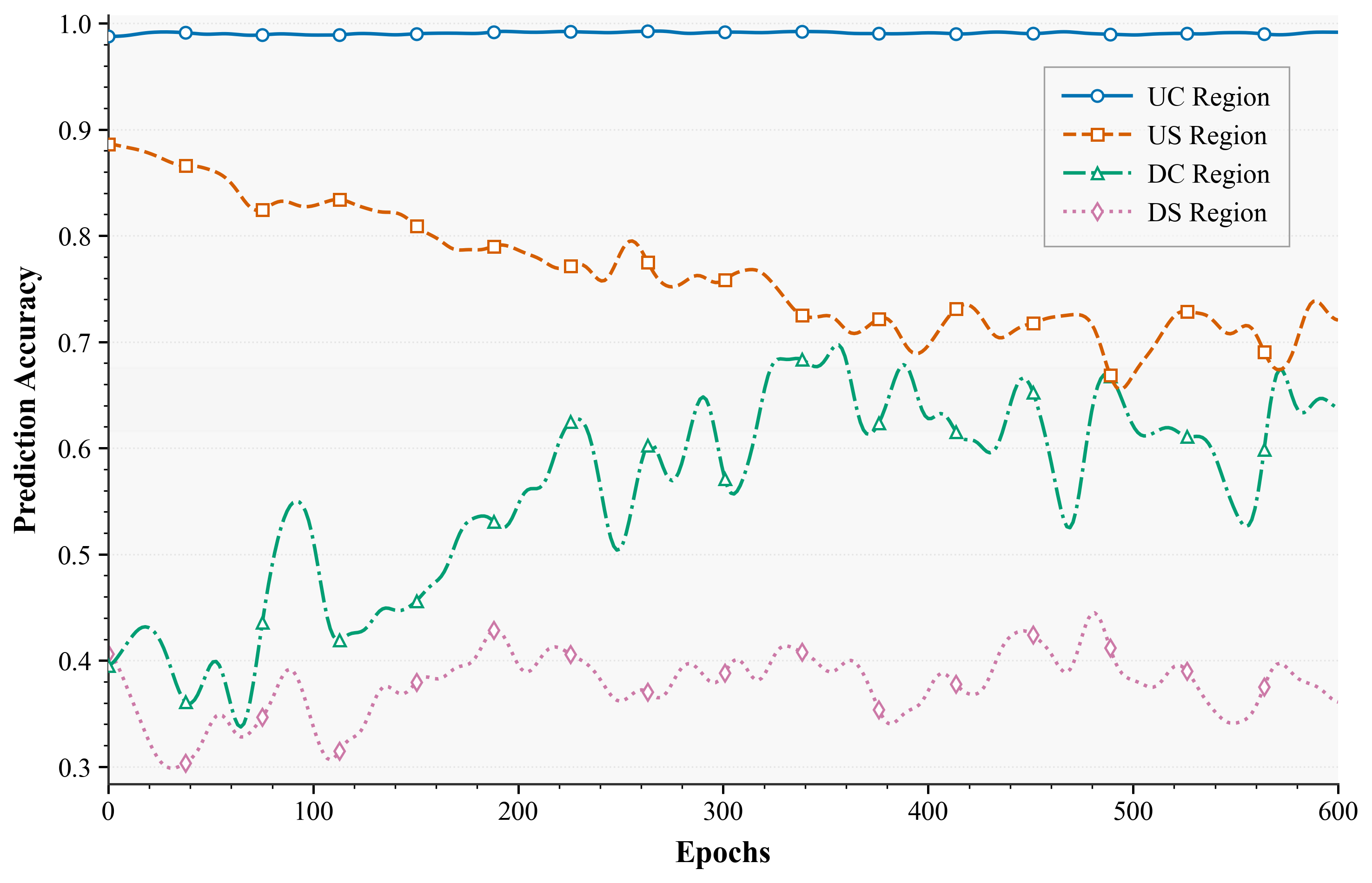}
    \caption{{The accuracy of four different regions (UC, US, DC, DS) in the CPS~\cite{chen2021semi} framework on the ACDC dataset at different training epochs.}}
    \label{fig:Acc_compare}
\end{figure}

\subsubsection{Analysis of different prediction confidence methods}
{To evaluate the effect of uncertainty quantification method on the performance, 
we considered several variants of uncertainty measurements for comparison: 1) Mean Absolute Deviation (MAD)~\cite{zhang2014thresholding}, 2) Variance~\cite{you2023rethinking}, 3) Standard Deviation (STD)~\cite{thamilarasi2019automatic}, and 4) Target-class probability~\cite{sohn2020fixmatch}.  As shown in \autoref{tab:Uncertainty}, for  both annotation ratios on the ACDC dataset, the target-class probability used in our method obtained the best performance. For example, when the annotation budget was a single volume, our method obtained an average DSC of 83.17\%, which is much better than MAD (78.58\%), Variance (80.36\%) and STD (81.41\%).} 

\subsubsection{Accuracy of Different Regions in Quadripartition}
{To demonstrate the rationale of our weight ranking for unlabeled images ($ w_{UC} > w_{US} > w_{DC} > w_{DS}$), we analyze the} {accuracy of these four regions under the CPS framework at different training epochs. As shown in \autoref{fig:Acc_compare}, UC achieves the highest accuracy, followed by US, while DC and DS exhibit progressively lower performance. It indicates that regions with more confident and consistent predictions are inherently more reliable than those with higher uncertainty or prediction discrepancy, which supports the weighting strategy of our quadripartition-based spatially heterogeneous loss.}

\begin{table}
  \centering
  \caption{{Comparison of different uncertainty estimation methods in the ACDC dataset.}}
    \scalebox{0.9}{\begin{tabular}{c|cc|cc}
    \toprule
    \multirow{2}[4]{*}{{Methods}}  &  \multicolumn{2}{c|}{{1 annotated volume}} & \multicolumn{2}{c}{{3 annotated volumes}} \\
\cmidrule{2-5}          &  \multicolumn{1}{c|}{{DSC(\%) \(\uparrow\)}} & {HD$_{95}$(mm) \(\downarrow\)} & \multicolumn{1}{c|}{{DSC(\%) \(\uparrow\)}} & {HD$_{95}$(mm) \(\downarrow\)} \\
    \midrule
    {MAD~\cite{zhang2014thresholding}}    & \multicolumn{1}{c|}{{78.58±13.06}} & {5.23±9.15} & \multicolumn{1}{c|}{{83.79±11.80}} & {2.78±4.47} \\
     {Variance~\cite{you2023rethinking}}    & \multicolumn{1}{c|}{{80.36±10.91}} & {4.43±7.56} & \multicolumn{1}{c|}{{85.76±8.49}} & {3.55±6.23} \\
     {STD~\cite{thamilarasi2019automatic}}    & \multicolumn{1}{c|}{{81.41±13.16}} & {4.56±7.88} & \multicolumn{1}{c|}{{86.64±8.13}} & {2.18±2.40} \\
    {Confidence~\cite{sohn2020fixmatch}} & \multicolumn{1}{c|}{\textbf{{83.17±9.69}}} & \textbf{{2.77±3.21}} & \multicolumn{1}{c|}{\textbf{{88.31±5.79}}} & \textbf{{1.66±1.80}} \\
    \bottomrule
    \end{tabular}}%
  \label{tab:Uncertainty}%
\end{table}%

\section{DISCUSSION}
\begin{figure}
    \centering
    \includegraphics[width=1\linewidth]{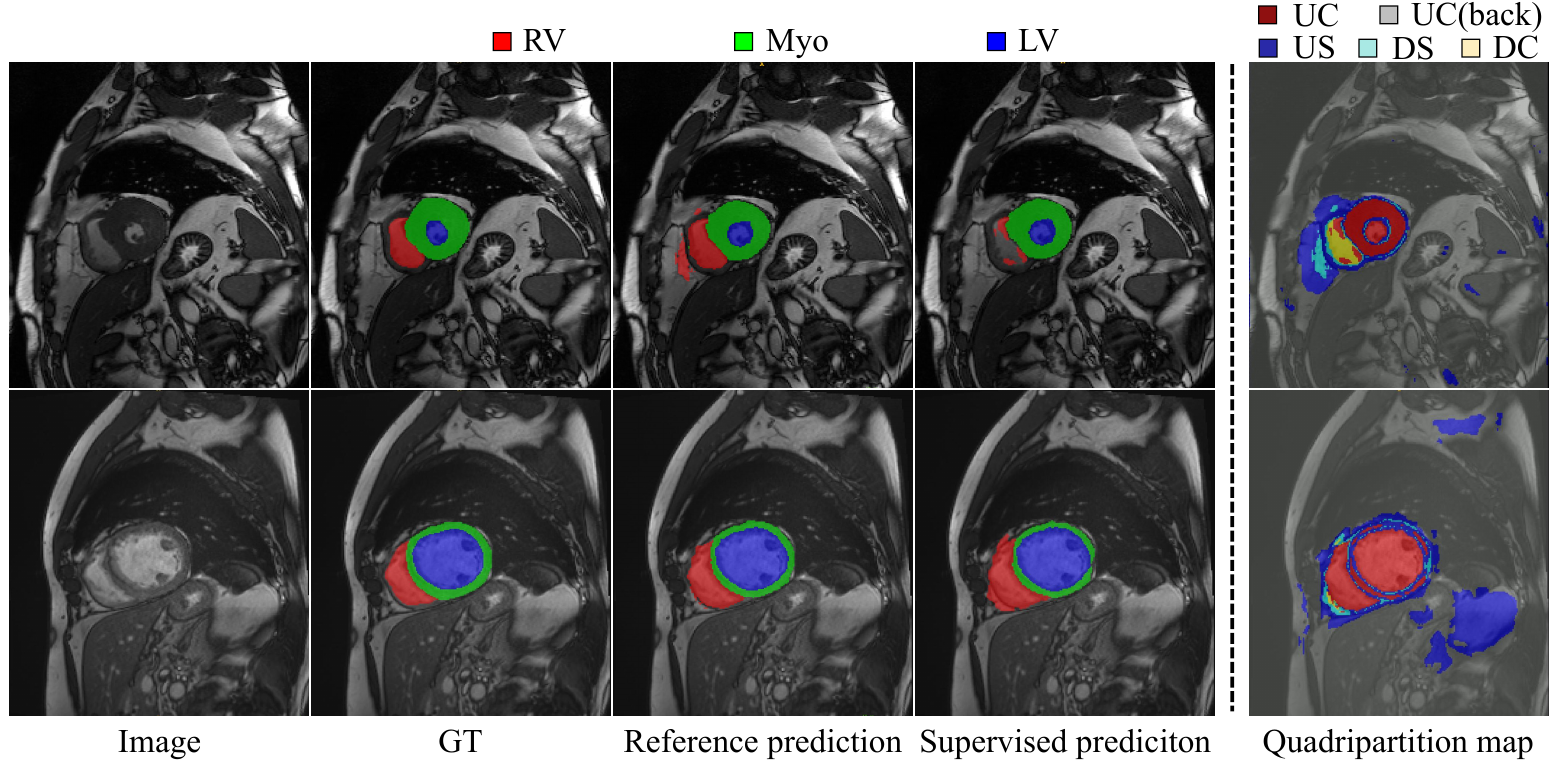}
    \caption{Visualization of the quadripartition of pseudo labels on the ACDC dataset.}
    \label{fig:foul-label}
\end{figure}

Semi-supervised medical image segmentation is challenged by unreliable supervision for unannotated images when only a small portion of the training set is labeled. Existing methods mainly focus on different strategies for generating supervision signal (i.e., reference prediction for consistency regularization or pseudo-labels) based on perturbations on input, feature or network structures~\cite{tarvainen2017mean,sohn2020fixmatch,ouali2020semi,chen2021semi,liang2021r}, and have ignored potential noise in labeled images. In this work, we propose a general loss function that simultaneously leverage consistency and uncertainty to better mine the rich information of reference predictions for learning.  It is a unified framework for dealing with both labeled and unlabeled images based on a quadripartition of the reference prediction into UC, US, DC and DS. \autoref{fig:foul-label} shows a visualization of pseudo-labels and the quadripartition maps on the ACDC dataset. In the first row, the pseudo-label has a high prediction accuracy for the left ventricle and myocardium, and the right ventricle has an under-segmentation. The   quadripartition map shows that the left ventricle and myocardium aligned well with the UC region, while the US region is mainly around border of the target. DC and DS are mainly in the under-segmented region and the incorrect pseudo-label, respectively. Assigning a  higher weight to DC than DS can effectively mine the reliable part of pseudo-labels for supervision while suppressing the effect of incorrect/noisy part. A similar  observation can be  found in the second row. {In \autoref{fig:view1_new}, we further show the difference of the quadripartition maps between CPS and CPS + MetaSSL. By using MetaSSL, the prediction is better, and the UC region is larger, i.e., more reliable pseudo-labels are generated, which further contributes to higher performance after more training epochs. In addition, on the boundary part, the reference prediction is usually less reliable due to ambiguity. The US region of MetaSSL matches well with the boundary, showing its effectiveness in dealing with noise, as shown by the quadripartition maps in the 300 and 600 epochs.}

\begin{figure}
    \centering
    \includegraphics[width=1\linewidth]{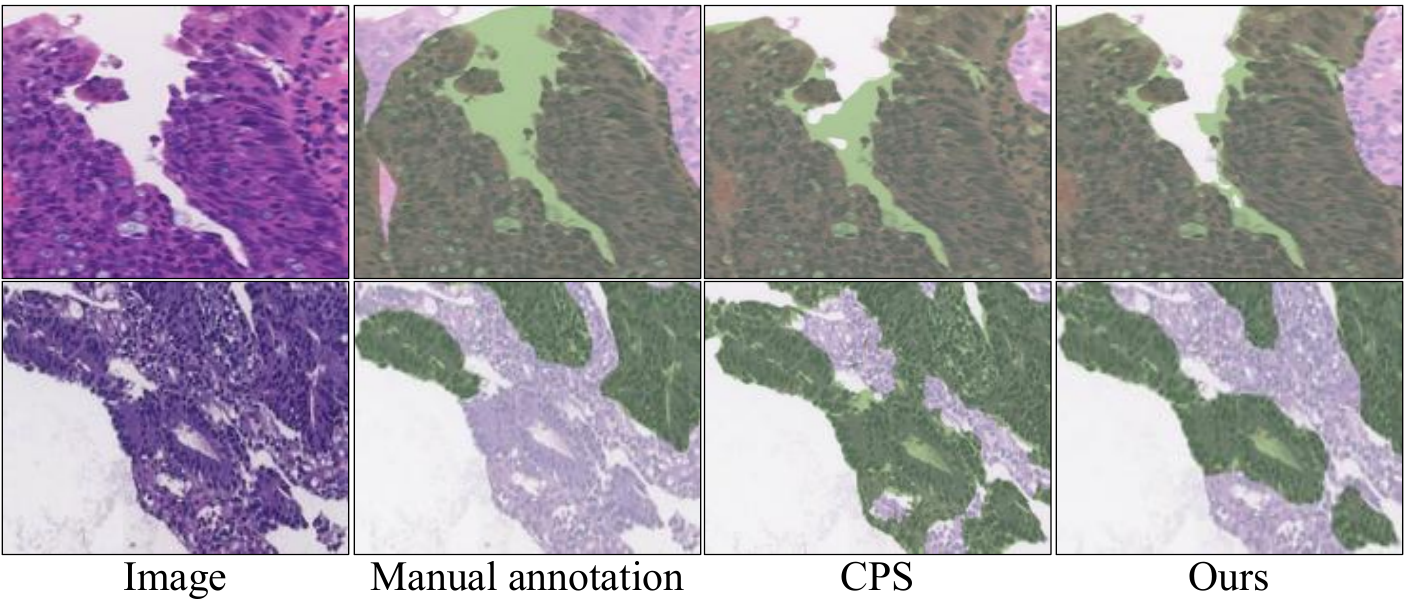}
    \caption{Examples of dealing with noisy annotations on labeled images in  the DigestPath dataset.}
    \label{fig:path_com}
\end{figure}

Our spatially heterogeneous loss is also effective in dealing with noisy annotations on the labeled images. \autoref{fig:path_com} shows the predictions of our method on labeled images on the DigestPath dataset. Note that the annotation contains some noise, i.e., over-segmentation and under-segmentation are observed in the two cases, respectively. The predictions of our model effectively corrected the noise, showing noise-robust learning ability of our method. In contrast, the CPS model~\cite{chen2021semi} is more affected by the noisy annotation in the labeled images.



{Results in \autoref{tab:sota_ACDC}, \autoref{tab:sota_path} and \autoref{tab:sota_LA} show that our method leads to significant improvement for most of the different SSL backbones, showing the generalizability of our MetaSSL. Note that on the DigestPath dataset, the improvement when applied to BCP~\cite{bai2023bidirectional} is relatively small. This is mainly  due to two reasons. First, the amount of annotation on the DigestPath dataset is not very small, i.e., 5 WSIs and 10 WSIs, which corresponds to 481 and 962 annotated patches, respectively. Such amount of annotations leads the room of improvement for top existing methods like BCP to be quite limited.  Second, the tumor region on the DigestPath dataset is diffusive with irregular shapes and sizes, making the segmentation more challenging than the organ segmentation tasks. For example, the upper bound of Fully Supervised Learning (FSL) is only 60.71\% in terms of Jaccard Index. Under an annotation ratio of 10\%, BCP and MetaSSL + BCP obtained a Jaccard Index of 56.98\% and 58.04\%, respectively. Considering the relatively small gap between BCP and FSL, the improvement brought by MetaSSL is signficiant, as it reduced the gap to FSL by 28.4\%, i.e., from 3.73 percentage points to 2.67 percentage points. Therefore, our method is still meaningful in terms of improvement. Note that that effectiveness of our loss function may be impacted by the dataset characteristics and annotation budget, and it is more advantageous for low annotation budget and relatively poor backbone SSL methods.}


{This work also has some limitations that could be addressed in the future. First, our loss function was mainly validated for CNN-baed SSL methods. In recent years, several methods tried to use novel network architectures such as Transformers for SSL. As our method is architecture-agnostic, it can be applied to such methods as well. Second, the weight of each region require tuning some hyper-parameters. Though we have tried to reduce the  hyper-parameters as much as possible in Section~\ref{sec:heterogeneus_loss}, the method still requires some efforts for tuning them. It is of interest to design some adaptive weighting methods to avoid this. }
\section{CONCLUSION}
In conclusion, we introduce a general semi-supervised medical image segmentation method MetaSSL based on quadripartition-based spatial heterogeneous loss functions.  For a multi-prediction-based SSL architecture, it leverages both consistency and confidence information to partition pixels into four categories to  highlight valuable ones and suppress noisy or less useful ones. The spatial heterogeneous loss functions not only improves the effectiveness of learning from pseudo-labels on unlabeled images, but also alleviate the effect of annotation noise on labeled images. The experiment showed that MetaSSL led to large performance improvement for several existing SSL frameworks, demonstrating its generalizability across different backbone SSL methods. 
In the future, it is of interest to explore the application of our method to 3D medical image segmentation tasks and deal with the distribution shift between labeled and unlabeled images. 

\bibliographystyle{IEEEbib}

\bibliography{ref}

\end{document}